\documentclass[10pt,twocolumn,letterpaper]{article} 

\usepackage{avss}
\usepackage{times}
\usepackage{epsfig}
\usepackage{graphicx}
\usepackage{amsmath}
\usepackage{amssymb}

\usepackage{fancyhdr}  

\avssfinalcopy 

\def\avssPaperID{50} 

\ifavssfinal\fi
\newcommand{\comment}[1]{}
\newcommand{\cb}[1]{#1}
\newcommand{\cbb}[1]{#1}
\newcommand{\cnn}[1]{#1}
\newcommand{\cnnn}[1]{#1}
\newcommand{\crr}[1]{#1}
\newcommand{\crrr}[1]{#1}

\usepackage[pagebackref=true,breaklinks=true,letterpaper=true,colorlinks,bookmarks=false]{hyperref}

\begin{document}

\title{The Multi-Strand  Graph for a PTZ Tracker}

\author{Shachaf Melman \\
Tel Aviv University\\
{\tt\small shachafme@gmail.com}
\and
Yael Moses\\
The Interdisciplinary Center\\
{\tt\small yael@idc.ac.il}
\and
G\'{e}rard Medioni ~~~~~~ Yinghao Cai\\
University of Southern California\\
{\tt\small medioni@usc.edu,caiyinghao@gmail.com}
}

\maketitle
\thispagestyle{empty}

\begin{abstract}
High-resolution images  can be used  to resolve  matching ambiguities  between trajectory fragments (tracklets),  which is  one of the main challenges in multiple target tracking. A PTZ camera,  which can pan, tilt and zoom, is a powerful and efficient tool that offers both close-up views  and wide area coverage on demand.  The wide-area view makes it possible to track  many targets  while the  close-up view allows individuals to be identified from high-resolution images of their faces.  A central component of a PTZ tracking system is a scheduling algorithm that  determines  which target to zoom in on. 

In this paper we study this scheduling problem from a theoretical  perspective, where the high resolution images are also used for tracklet matching. We propose a novel data structure,  the Multi-Strand Tracking Graph (MSG), which  represents the set of tracklets computed by a tracker and the possible associations between them. The MSG  allows efficient scheduling as well as resolving -- directly or by elimination -- matching ambiguities between tracklets. 
The main feature of the MSG is the auxiliary data saved in each vertex, which allows  efficient computation 
while avoiding time-consuming  graph traversal. Synthetic data simulations are used to evaluate our scheduling algorithm and to demonstrate its superiority over a na\"{\i}ve one.

\end{abstract}

\section{Introduction}

\comment{
\begin{figure}[b!]
\begin{tabular}{cc}
 \includegraphics[width=0.5\columnwidth]{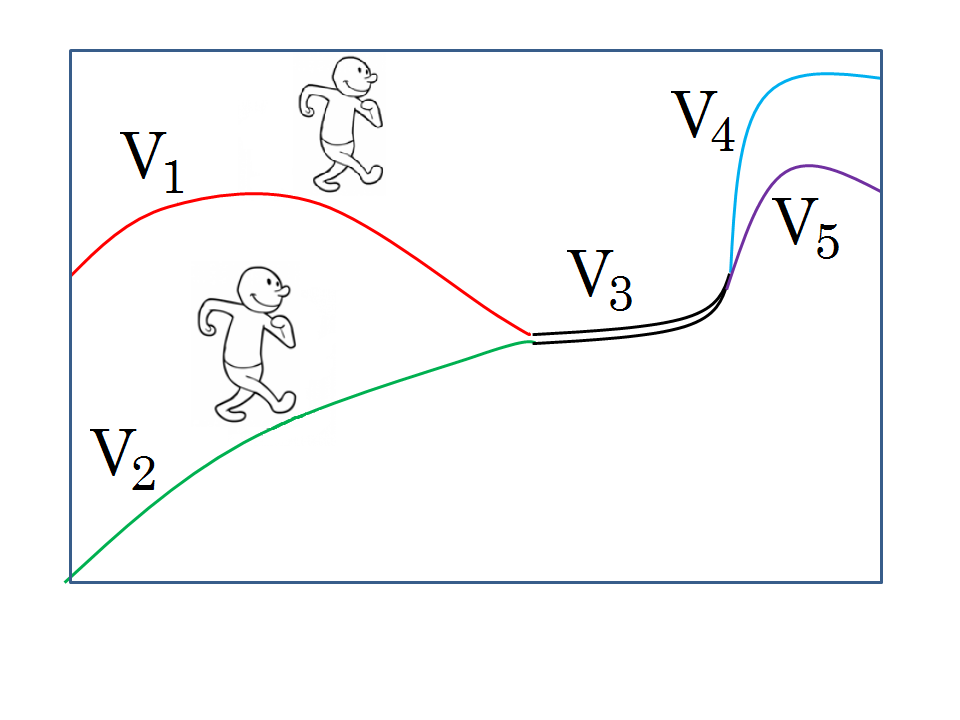} 	 &
 \includegraphics[width=0.5\columnwidth]{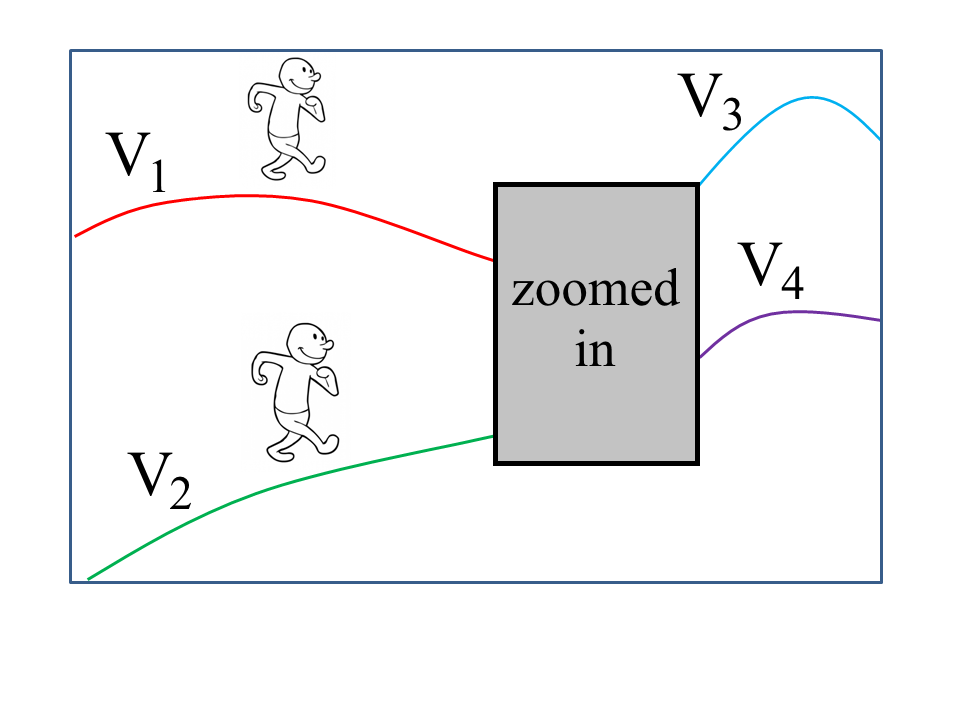} \vspace{-1cm}\\
~~~~~~~~~~~~~~~~~~~~~\includegraphics[width=0.15\columnwidth]{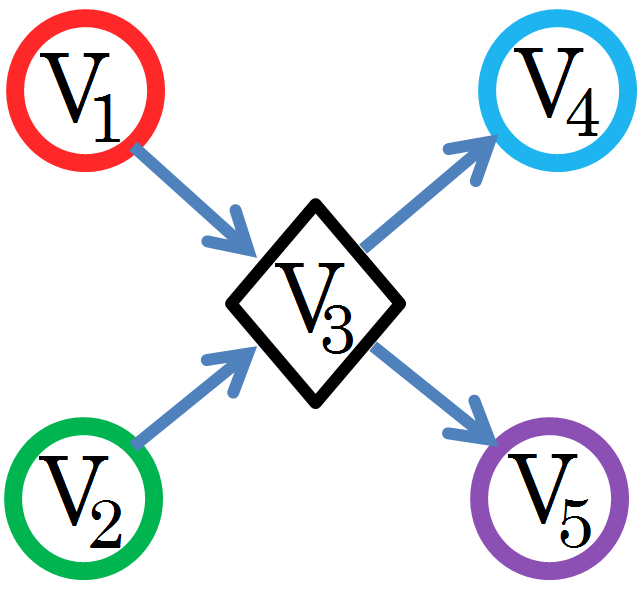} &	
~~~~~~~~~~~~~~~~~~~~~  	  \includegraphics[width=0.15\columnwidth]{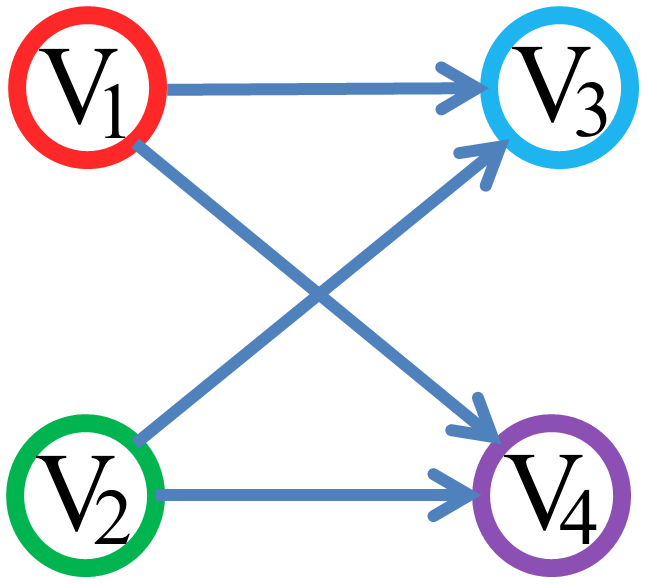} 	 \\

			(a) & (b) 
	\end{tabular}
\caption[]{\small  (a) Two targets  walk separately, join and then split. (b) Blind gap:  targets who walk separately  but  move out of sight when the camera zooms in on another target, and then  become visible again in the next zoom-out mode.
The graphs: circle nodes \cbb{represent} solo vertices; diamond \cbb{nodes} represent compound  \cbb{vertices}.
}
\label{fig:pairTargets}%
\end{figure}
}

We consider a system consisting of a single PTZ camera (which can pan,  tilt and zoom) to solve the problem of tracking multiple pedestrians while also capturing their faces.
A necessary component of such a system is a  scheduling algorithm that  determines at any \cbb{time step} whether to remain in  zoom-out mode or  to zoom in on a face. 
This paper presents an efficient new data structure, the multi-strand  graph (MSG), for multiple target tracking using  a single PTZ  system, and a novel scheduling algorithm based on it. 

\comment{
\begin{figure*}[t!]%
\centering
\begin{tabular}{ccccc}
\includegraphics[width=0.544\columnwidth]{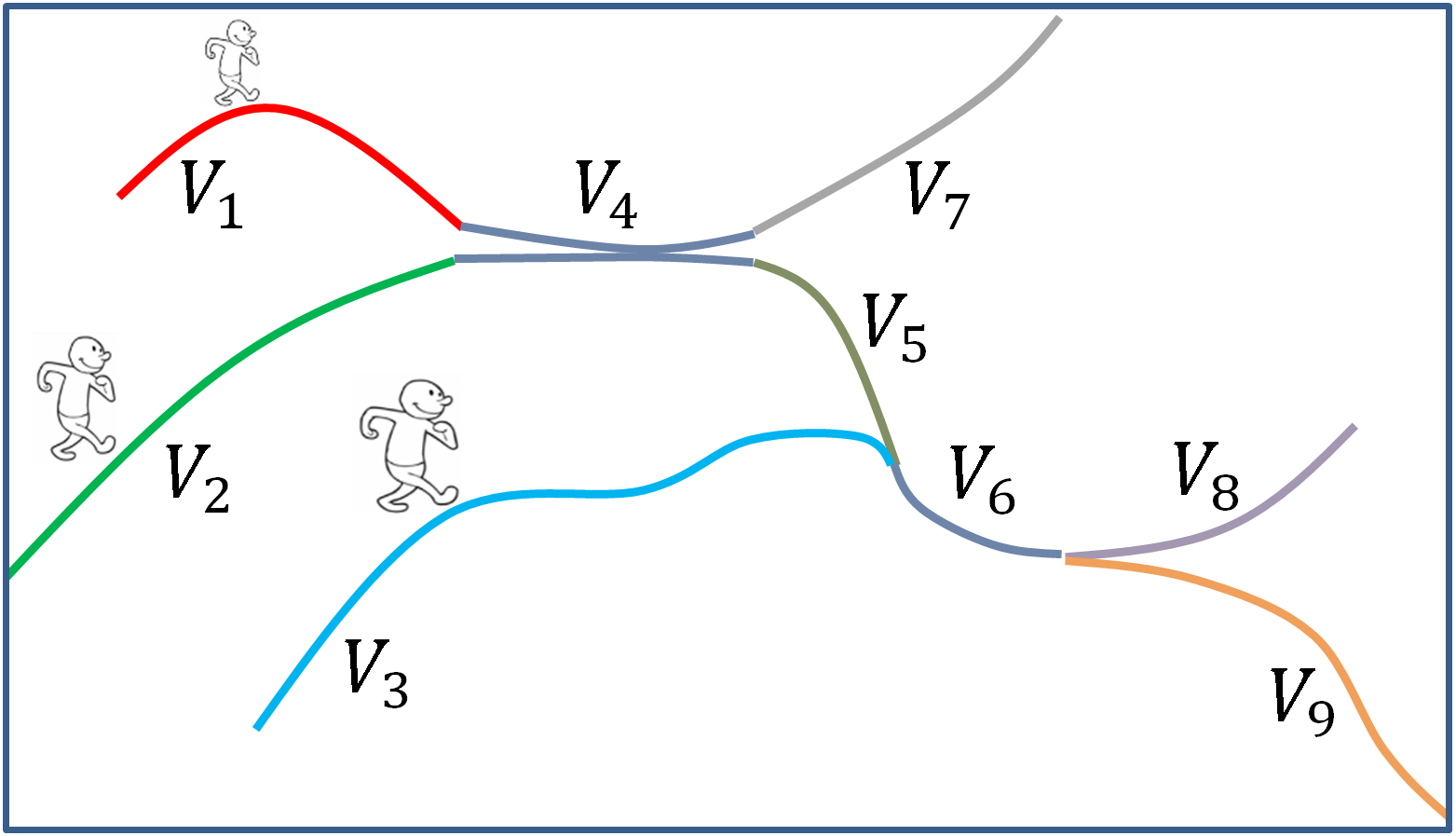} &
\hspace{-0.2cm}\includegraphics[width=.344\columnwidth]{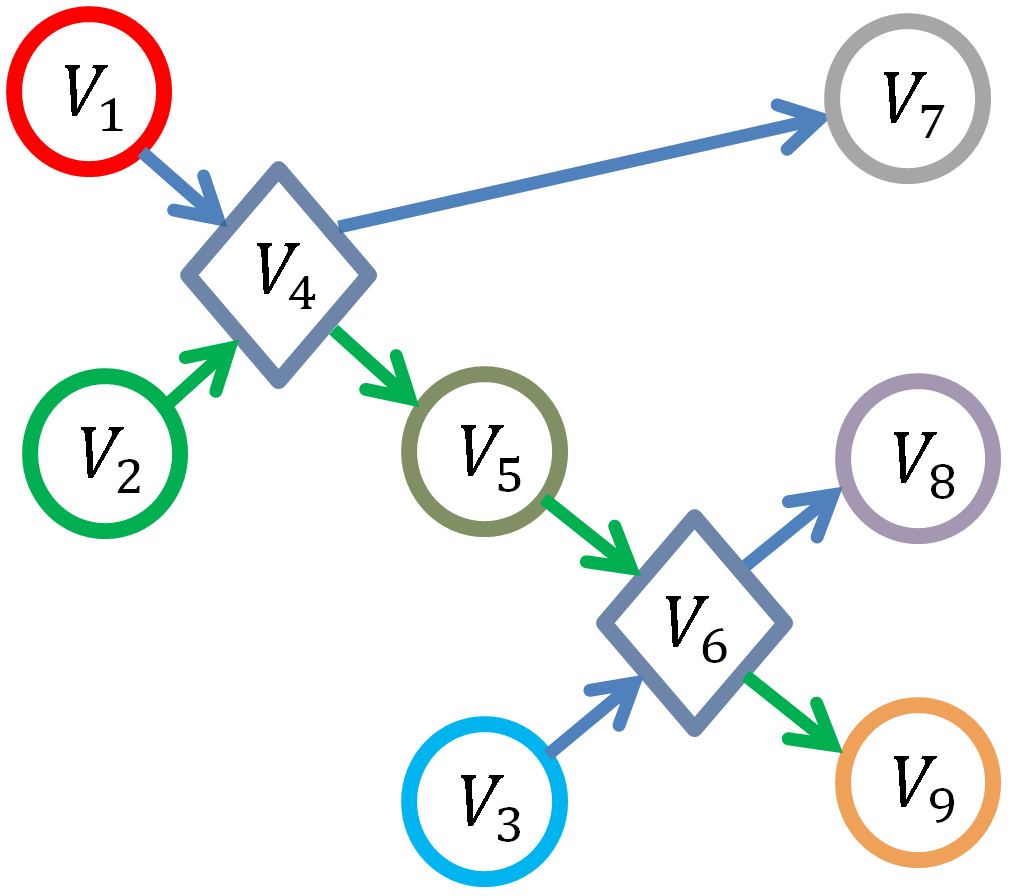} &
\hspace{-0.3cm} \includegraphics[width=.456\columnwidth]{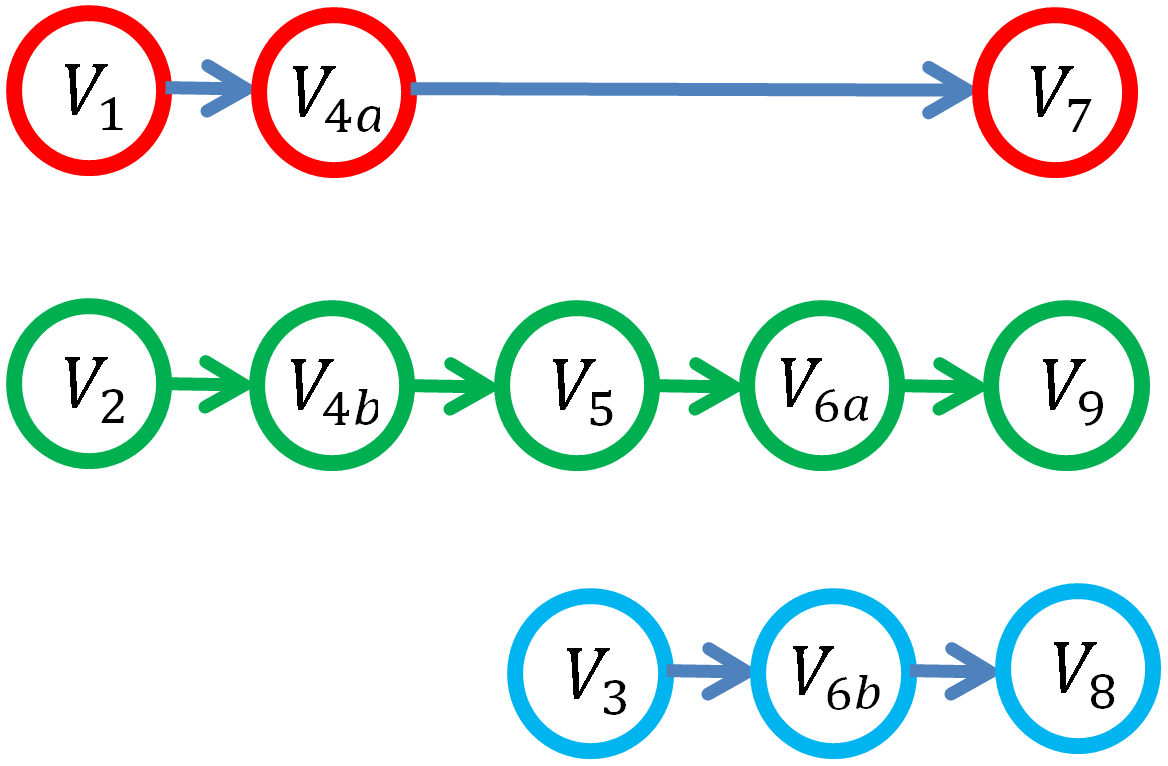} &
\hspace{-0.3cm} \includegraphics[width=.165\columnwidth]{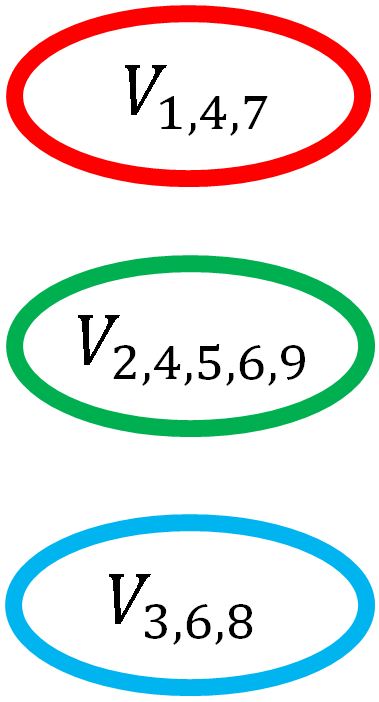} &
\hspace{-0.4cm}
\includegraphics[width=.491\columnwidth]{Figures/TRs2.png} \vspace{-1.1cm}\\
 &
 & 
 &
 &
~~~~~~~~~~~~~~~~~~~~~  	  \includegraphics[width=0.15\columnwidth]{Figures/Graph2.png} 	 \\ 
(a) & (b) & (c) & (d) & (e)
\end{tabular}
\caption{(a) Three targets walk in a scene, with two join and split events. (b) The corresponding MSG. \cbb{Circle nodes represent solo vertices, and diamond nodes -- compound vertices.} (c) First untangling step after matching $v_9$ to $v_2$ using high-resolution images. (d) The final MSG after untangling: \cb{each solo vertex represents a full target trajectory.} \cbb{(e) A blind-gap scene and its MSG: targets who walk separately but move out of sight when the camera zooms in on another target, and then become visible again in the next zoom-out mode. }}%
\label{fig:ThreeTargets}%
\end{figure*}
}

\begin{figure}[b!]
\begin{tabular}{cc} \hspace{-0.5cm}
 \includegraphics[width=0.5\columnwidth]{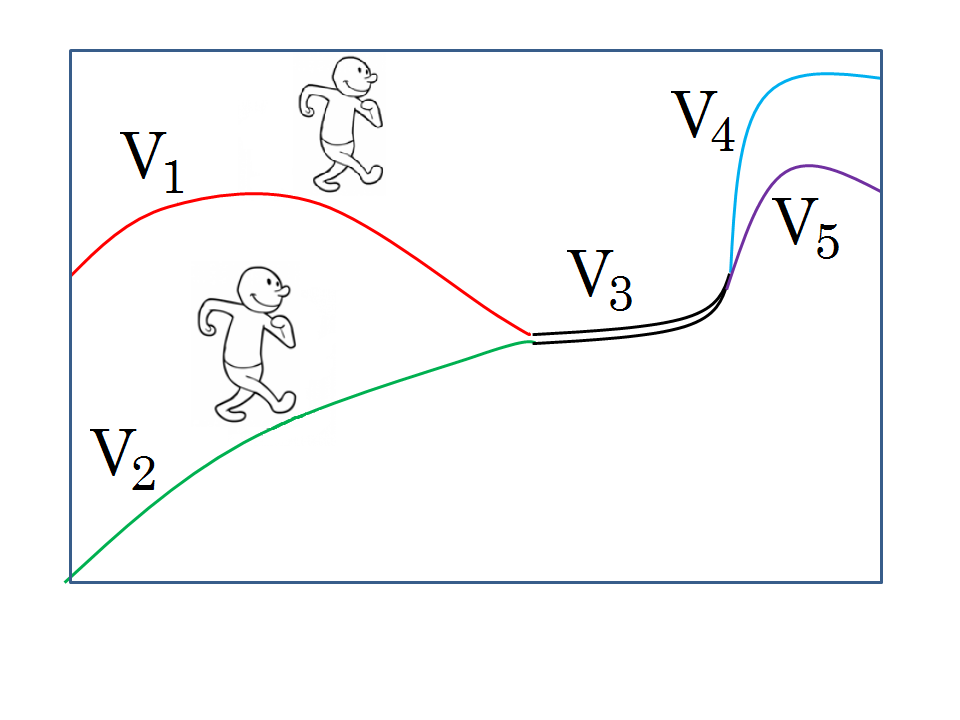} 	 &
 \includegraphics[width=0.5\columnwidth]{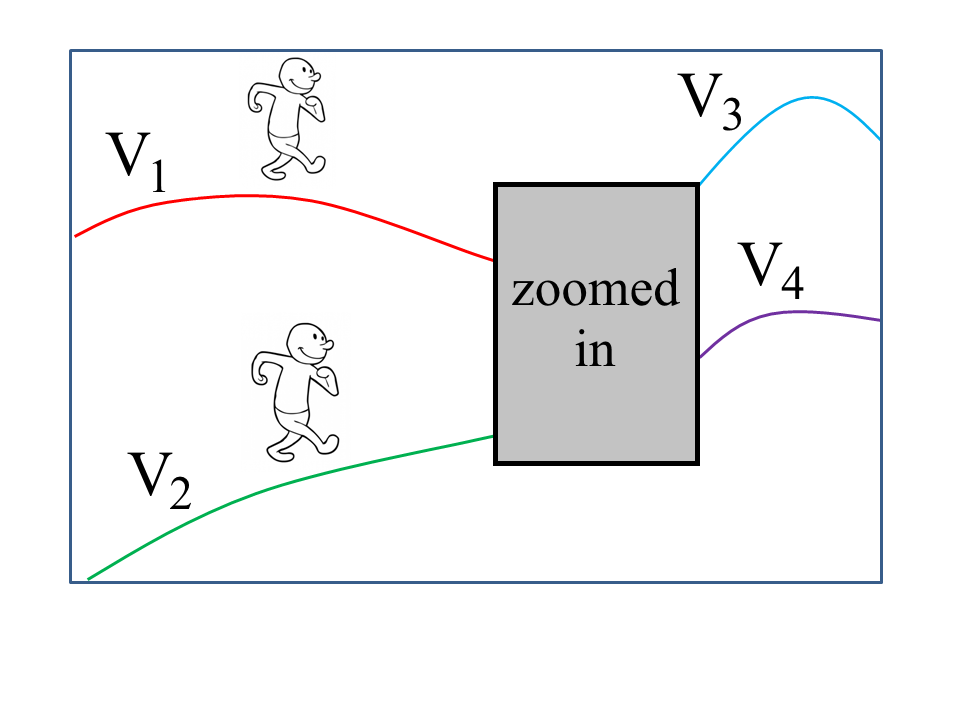} \vspace{-1.1cm}\\ \hspace{-0.5cm}
~~~~~~~~~~~~~~~~~~~\includegraphics[width=0.2\columnwidth]{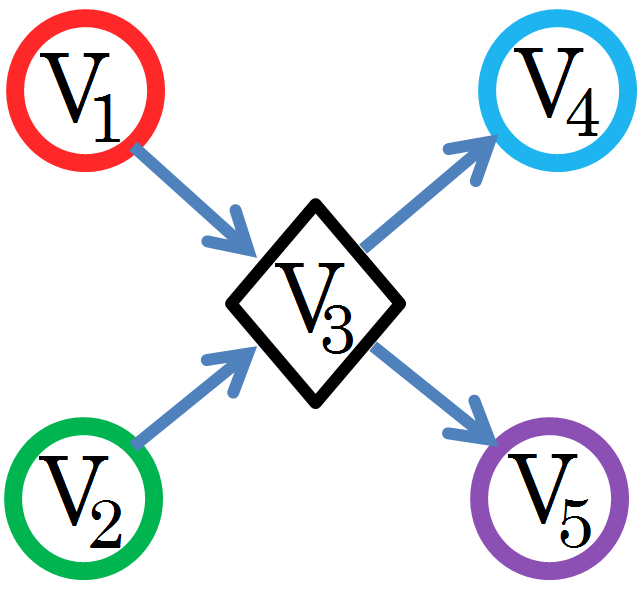} &	
~~~~~~~~~~~~~~~~~  	  \includegraphics[width=0.2\columnwidth]{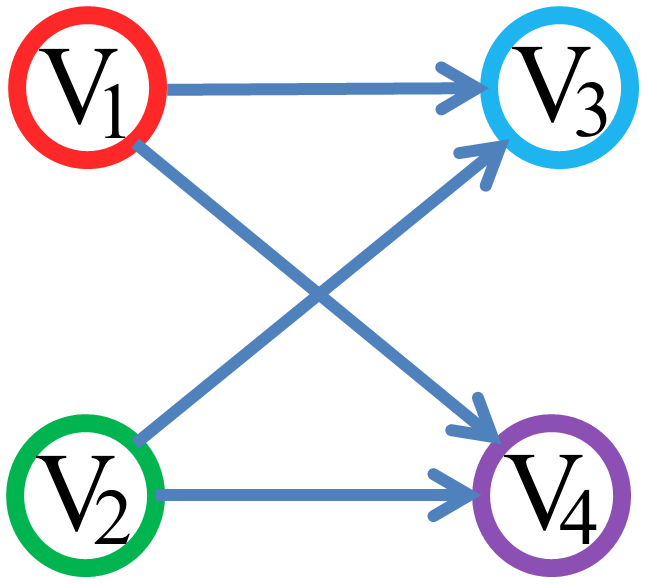} 	 \\ \hspace{-0.5cm}

			(a) & (b) 
	\end{tabular}
\caption[]{\small  (a) Two targets  walk separately, join and then split. (b) A blind-gap scene and its MSG: targets who walk separately  but  move out of sight when the camera zooms in on another target, and then  become visible again in the next zoom-out mode.
Circle nodes represent solo vertices. A diamond node represents a compound  vertex.}
\label{fig:pairTargets}%
\end{figure}

\begin{figure*}[t!]
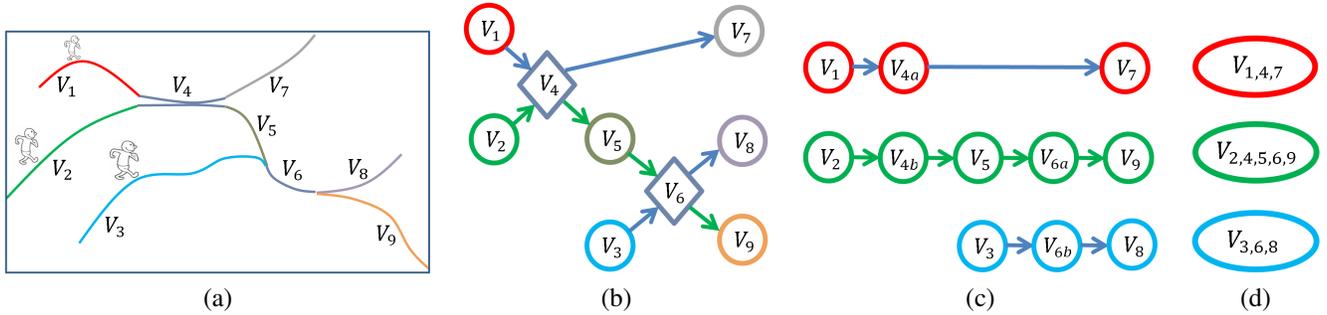
%
\centering
\begin{tabular}{cccc}
\includegraphics[width=0.68\columnwidth]{Figures/ThreeTargets2} &
\includegraphics[width=.49\columnwidth]{Figures/figure2a} &
\includegraphics[width=.57\columnwidth]{Figures/figure2b} &
\includegraphics[width=.207\columnwidth]{Figures/figure2c} \\
(a) & (b) & (c) & (d)
\end{tabular}
\caption{(a) Three targets walk in a scene, with two join and split events. (b) The corresponding MSTG. (c) First untangling step after matching $v_9$ to $v_2$ using high-resolution images; (d) The final MSTG after untangling: \cb{each solo vertex represents a full target trajectory. Circle nodes represent solo vertices, and diamond nodes -- compound vertices.}}%
\label{fig:ThreeTargets}%
\end{figure*}

Our method aims to overcome one of the main challenges of a multiple target tracker:  trajectory fragmentation. Such fragmentation is caused by occlusions, by the joining of two or more targets  who then split (e.g.,  Figures~\ref{fig:pairTargets}(a),~\ref{fig:ThreeTargets}(a)), or when the PTZ  camera zooms in on another target, creating what we  call a  {\em blind gap} (Figure~\ref{fig:pairTargets}(b)).  Matching trajectory fragments (tracklets) is complicated due to  ambiguities caused  by similarity in  appearance and location of  different targets. 
We propose using the faces captured in zoom-in mode together with the available information of the system state,
to  resolve such ambiguities.  

The  objective of the proposed system is to maximize the total length of the {\em labeled  tracklets}, that is, trajectory fragments with associated high resolution face images captured in zoom-in mode.  Formally, let $Z$ be the set of targets, and let 
$\{\tau_i (z)\}_{i=1}^{n_z}$ 
be the set of its detected tracklets.  
For each target \cbb{$z\in Z$ we define} $\tau_{L}(z)$ to be the union of the labeled \cb{tracklets} of $z$. The objective is given by
\begin{equation} 
 M=\tfrac{\sum_{z\in Z}{|\tau_L(z)|}}{\sum_{z\in Z}\sum_{i=1}^{n_z}|\tau_i (z)|},
 \label{eq:M}
\end{equation}
where $|.|$ denotes the tracklet length.
Our scheduler selects the target with the highest probability to maximize $M$ to be the next target that will be zoomed in on. Note that resolving ambiguities (e.g., matching $v_2$ and $v_9$ in Figure~\ref{fig:ThreeTargets}(a)) can greatly increase $M$.

We introduce the Multi-Strand Tracking Graph (MSG), which represents the tracklets computed by a  tracker and  their possible associations  (\cnn{its basic structure is }similar to \cite{sullivan2006a}). We show that a straightforward use of the graph for the abovementioned task requires a graph traversal. The main contribution of this paper is \cnn{the proposed} auxiliary data stored in each vertex. We use this data to  efficiently compute the system state information  without traversing the graph. The graph is constructed online and the auxiliary data is recursively computed  based  only on the vertex itself and \cbb{on} its direct parents. Hence, all the required information is available when scheduling decisions are made. \cbb{Other contributions} of this paper \cbb{are} the use of high-resolution images to resolve \cbb{matching ambiguities of} tracklets and the design of an efficient scheduling algorithm
\cbb{that uses} the MSG.

\begin{figure}[t!]
\centering
\begin{tabular}{cc}\hspace{-0.6cm}
\begin{tabular}{c}
\includegraphics[width=0.5\columnwidth]{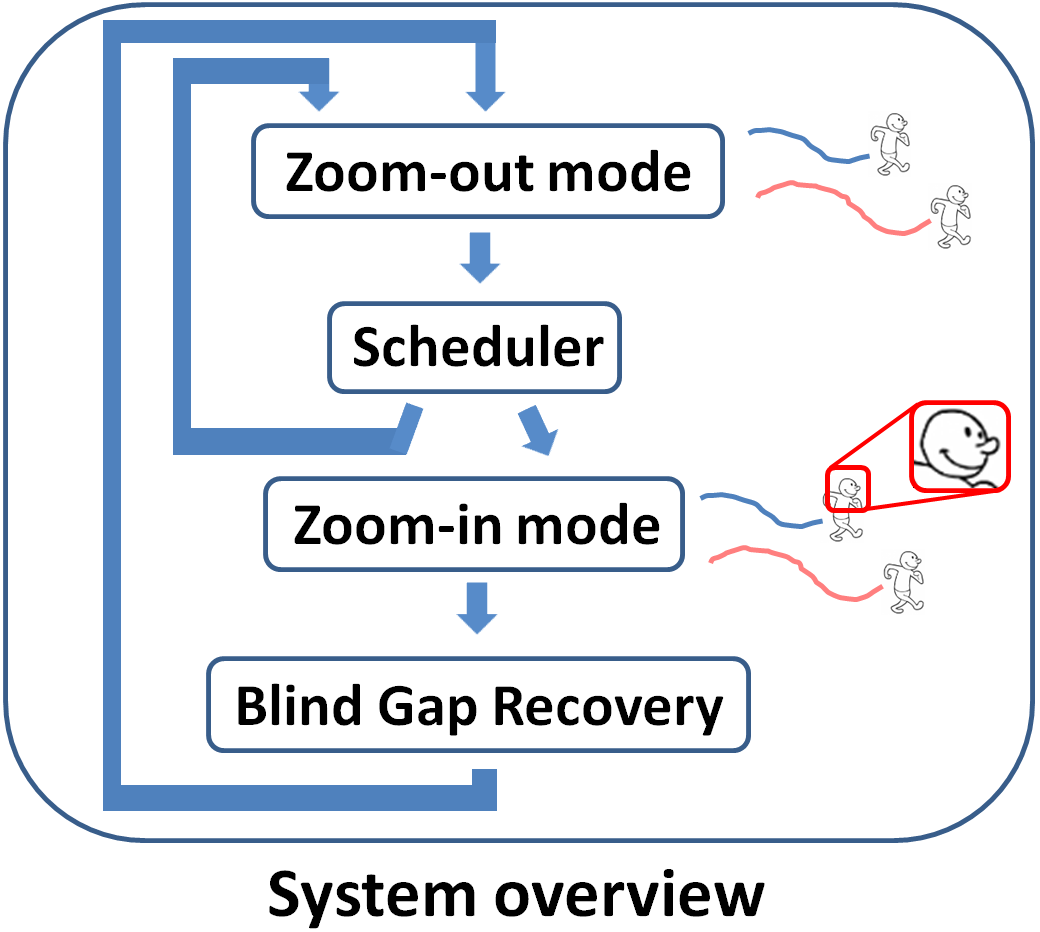} \\ 
\includegraphics[width=0.55\columnwidth]{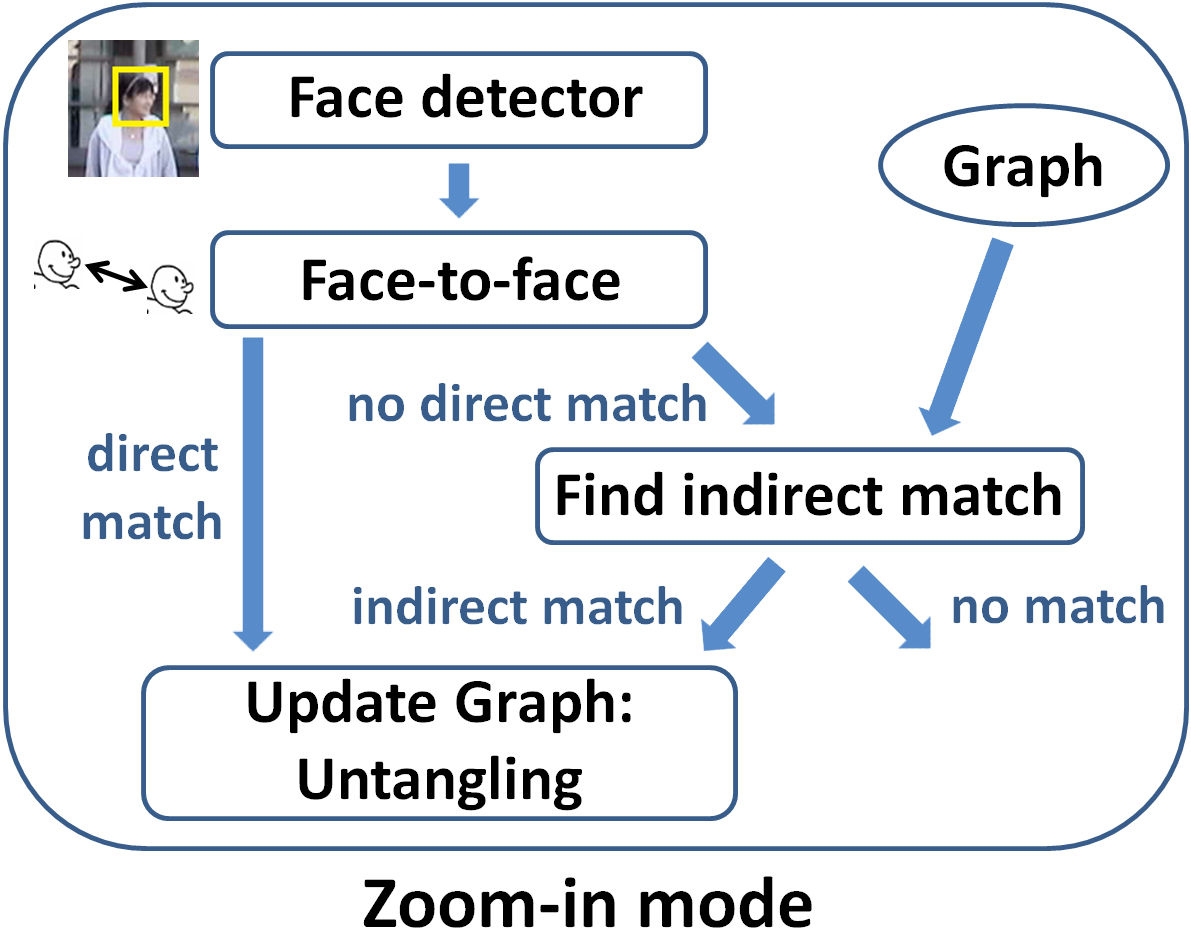} \\
\end{tabular} &
\hspace{-0.8cm}\begin{tabular}{c}
\includegraphics[width=0.45\columnwidth]{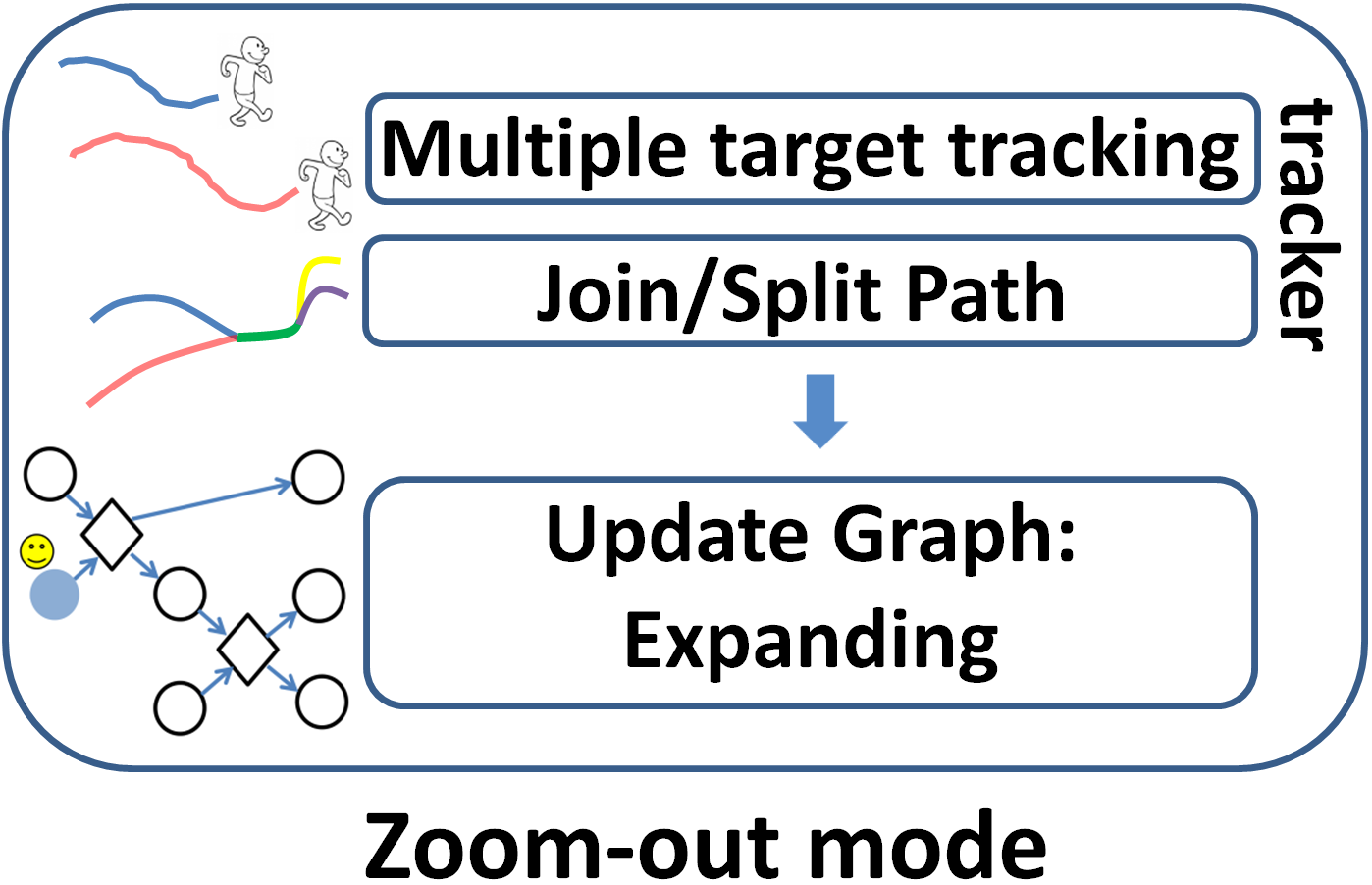} \\ \\
\includegraphics[width=0.45\columnwidth]{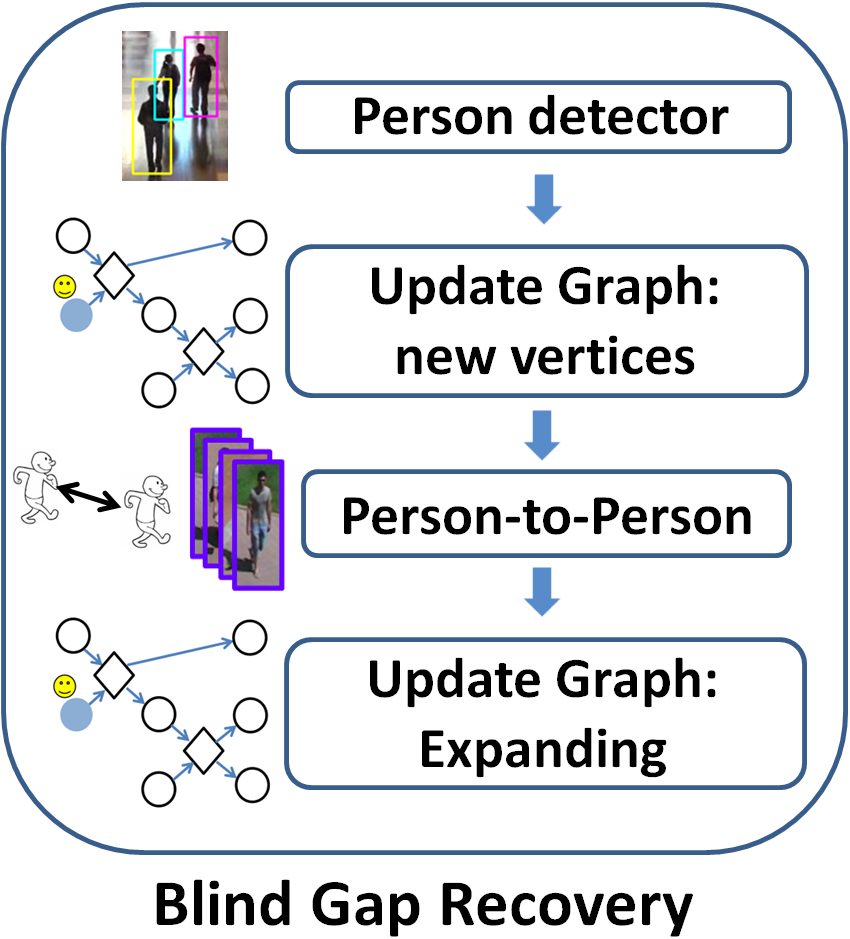} 
\end{tabular}
\end{tabular}
\caption[]{\small An overview of the system and its operation modes  }
\label{fig:overview}%
\end{figure}

 \vspace{0.2cm}
\noindent{\bf System overview:}
The tracking system considered in this paper consists of a single PTZ camera,  and several components, described below,  are assumed to be available.  These include a tracker that  detects and tracks pedestrians in zoom-out mode. It  also detects \cbb{joining and splitting events} of two or more  targets moving together (\cnn{as in}~\cite{sullivan2006a}). According to the proposed scheduler, the system selects a person to zoom in on using  a camera control algorithm. The control algorithm chooses  the  FOV that makes it possible to  zoom in on  selected target (e.g., \cite{cai2013}). In the \cbb{zoom-in} mode, a  face image is acquired and a face-to-face  and a face-to-person matchings are computed. The system then zooms out, to the same wide  view, to continue tracking. A person-to-person matching  module \cbb{associates tracklets}  when returning from  zoom-in mode  or after targets split from a group.  Figure~\ref{fig:overview} summarizes the system components. Our contribution to the system is the graph representation (MSG) and the efficient scheduling algorithm.

\section{Previous Work}
\label{sec:previousWork}
Scheduling of a single PTZ camera was considered in \cite{arambel2004,cai2013,salvagnini2014,sommerlade2008}.
Scenarios of joining/splitting targets  
 were considered in  \cite{arambel2004,cai2013}. The goal of
 \cite{arambel2004} was to  minimize the slew time of an aerial camera tracking cars.  
High-resolution images were used to remove incorrect prediction hypothesis (stored as a tree). 
The greedy policy in \cite{cai2013} aimed to maximize the number of captured faces, considering the predicted time of each target to exit the scene and its movement angle w.r.t. the camera.
An information-theoretic approach \cite{salvagnini2014,sommerlade2008} aims to decrease location uncertainty  while capturing high-resolution images.  
A distributed game-theoretic approach  for scheduling multiple PTZ cameras \cite{morye2014} aims to  maximize the targets' image quality and  to capture their  faces.
None of the above scheduling algorithms considered the goal of resolving tracklet-matching ambiguities.

Other systems considered  setups with  both fixed and PTZ cameras, in a master-slave configuration. 
Such setups are \cnn{less challenging since} \crr{a} fixed camera continuously views the entire region. They  vary from a single master and a single slave \cite{bagdanov2005,costello2004} to multiple masters and multiple slaves \cite{costello2005,lim2006,qureshi2007,ward2009}.  
The objectives  in these studies are to acquire once \cite{costello2005}, or as many times as possible \cite{costello2004}, the face of each target, or to minimize camera motion \cite{bagdanov2005}.  The scheduling methods consider the expected distance from the camera  \cite{qureshi2007}, the viewing angle \cite{bagdanov2005,lim2006,arambel2004,ward2009}, and expected occlusions \cite{lim2006,arambel2004}.
In addition to these objectives, our algorithm   also considers how zooming in contributes to the resolution of past and future ambiguities of tracklet matching.

Graphs were previously used  to  represent relations between tracklets \cbb{ \cite{henriques2011,prokaj2011,wang2014,Wu2011,yang2012}}, where the weighted edges reflect the appearance similarity and the consistency of location with respect to the computed motion direction and sometimes speed.  A graph with a similar structure to the MSG  \cite{henriques2011,Nillius2006,sullivan2006a}
was  used to associate isolated tracklets of targets  with indistinct appearance as well as tracklets of a set of targets that cannot be separated. The  \cbb{joins/splits} of targets \cbb{were} computed by a tracker. 
The association of single target tracklets is solved by  finding the most probable set of paths.
All these papers use  the target's location and only one appearance descriptor level for matching while we use both \cbb{low- and high-resolution} images. 
\cnn{Moreover, they do not use auxiliary data, which allows efficient scheduling and online graph updating in our method.}

\section{Method } 
\label{sec:method}
We first describe the basic structure of the MSG graph. 
\crr{Next,} we extend the MSG graph with \crr{auxiliary} data for efficient matching  by elimination. Finally, our  scheduling algorithm is presented.

\subsection{Graph Definition}
\label{sec:graphDefinition}
The basic structure of the  MSG  is a dynamic augmented graph,  $G=(V,E)$, where $V$  represents the set of tracklets computed by the tracker,  and $E$ the candidate associations of different tracklets computed by some available matching algorithm (similar to \cite{sullivan2006a}). Each vertex is associated with the information regarding  its tracklet.  
We consider two types of vertices that represent  two types of tracklets. A  \emph{solo vertex} represents the tracklet of a single target  and a \emph{compound vertex} represents the shared tracklet of \emph{joined} targets, that is, a set of targets that \cb{walk together} (Figure~\ref{fig:ThreeTargets}(b)).
 
A directed edge, $e=(v_i,v_j)\in E$, represents the case in which at least one of the targets associated with~$v_i$ may also be associated with~$v_j$, and $v_i$ and $v_j$  correspond to consecutive time intervals (ignoring zoom-in time).  
A compound vertex is generated as the child of other vertices when  the tracker detects that the targets' trajectories are joined into indistinguishable tracklets \cbb{(see Figures~\ref{fig:ThreeTargets}(a),~\ref{fig:ThreeTargets}(b)).} A new solo vertex is generated when a new target enters the scene,  a target trajectory splits from those of others (as a child of the compound vertex), or a target reappears   when the camera  returns to zoom-out (after a blind gap). 

Edges between  solo vertices are generated at \cbb{consecutive} layers (e.g., when returning from zoom-in mode), according to a matching algorithm that is based on the targets' low-resolution images captured in zoom-out mode and on their locations.
When the matching of a new solo vertex is ambiguous, edges are set \cbb{between} the vertex and  all the matching candidates, \cbb{forming an  
 \emph{X-type ambiguity}} (see Figure~\ref{fig:pairTargets}(b)). Additional edges are generated between compound and \cbb{solo vertices,} based on the tracker's detection of \cbb{splitting and joining} targets.

\subsection{Untangling}
When no ambiguities are present, the trajectory of each target can be fully recovered, 
\cnn{and the graph contains only unconnected solo vertices.} We wish to reduce as much as possible the number of vertices by concatenating consecutive tracklets to a single tracklet when possible. 
When  a  univocal matching exists for a \cbb{consecutive} set of tracklets (all of the  same target), their corresponding vertices form a {\em solo chain} in the graph -- \cbb{in which} the  \cbb{out degree} and the \cbb{in degree of a vertex's parent and child},  respectively,  are one (Figure~\ref{fig:ThreeTargets}(c)). A solo chain, of any length, can be merged to a single \cb{solo} vertex (Figure~\ref{fig:ThreeTargets}(d)). A compound chain can be \cbb{defined and} merged similarly.

A univocal matching of a pair of non-consecutive solo vertices, $\{u,v_L\}$,  can be computed using an available face-to-face matching algorithm. 	In addition, an  \emph{indirect match} can also be obtained by elimination
(see Section~\ref{sec:indirect}). 

Such  \crr{a} univocal matching of $\{u,v_L\}$  can be used \crr{for untangling} the graph as long as the connected component of $u$ and $v_L$ is a DAG (a graph that does not contain cycles).
In this case,  a breadth-first-search (BFS) algorithm is used  to recover the graph path, $\ell(u,v_L)$ (e.g.,  $\ell(v_2,v_9)$ in Figure~\ref{fig:ThreeTargets}(b)). 
	All vertices of $\ell(u,v_L)$ are guaranteed to represent consecutive tracklets of the same target.
Hence, edges `to' and `from'  solo vertices of $\ell(u,v_L)\setminus u$  (representing X-type ambiguities) are removed except those that are part of the path. Each compound vertex $v_{comp} \in \ell(u,v_L)$ is split into two vertices. One solo vertex represents only the labeled target and is linked only to the solo chain. The second vertex, $v_{split}$, represents the remaining targets of the compound vertex and is disconnected from the chain  (Figure~\ref{fig:ThreeTargets}(c)). As a result, a solo chain of the \cbb{labeled target, and possibly additional solo chains of other targets, are obtained. Each chain} can be merged into a single solo vertex (Figure~\ref{fig:ThreeTargets}(d)). 
\crr{Note that no information is lost in the untangling process.}

\subsection{Matching by Elimination}
\label{sec:indirect}
When there is sufficient confidence that a labeled vertex cannot be matched to any of the previously labeled  vertices, the vertex can \crrr{sometimes} be  indirectly matched to an unlabeled vertex by elimination.
For example, \cbb{assume that} $v_1, v_2$ and $v_9$ \cb{in Figure~\ref{fig:ThreeTargets}(b)} were labeled and no match  was found for the faces of either $v_1$ or $v_2$  with $v_9$. 
It is possible to  deduce 
that $v_3$ is the correct match to $v_9$. \cbb{Similarly, if only $v_3$ and $v_9$ were labeled, the non-source $v_5$ is deduced to be this match.} We next define when an  indirect match can be found in the general case, and how to compute it efficiently. 
Let  $V_L$ be the set of labeled solo vertices. We define an  {\em unlabeled path} between $w$ and  $v$, $\tilde \ell (w,v)$,  to be a path that does not contain any labeled vertex except   possibly $w$ and $v$, that is, \cbb{ $\forall u\in \ell(w,v)\setminus \{w,v\}, u\notin V_L$}.  

\vspace{0.2cm}\noindent{\bf Claim~1:~~}
Sufficient and necessary conditions for a solo vertex $w\notin V_L$ to be an indirect match to $v\in V_L$ are (i) $v$ cannot be matched to a previously labeled vertex; (ii) there exists an unlabeled path, $\tilde\ell(w,v)$; (iii)  if an unlabeled solo vertex $w'$  satisfies (ii) then $w'\in \tilde\ell(w,v)$.
\vspace{0.2cm}

\noindent{\bf Proof:~~}
We begin with proving that (i)-(iii) are necessary conditions. Assume $w$ is an indirect match of $v_L$. 
Then (i) must hold since  otherwise $v_L$ can be directly matched; (ii) must hold since otherwise
 either $\ell(w,v_L)$ does not exist and hence  no match between $w$ and $v_L$ is possible, or $\exists w'\in \ell(w,v_L)$, where $w'$ is a labeled solo vertex. However, an indirect match of $w$ and $v_L$ implies a match between all solo vertices $u\in \ell(w,v_L)$ and $v_L$. Hence, $v_L$ could be directly matched to $w'$, which contradicts condition (i).
Finally, (iii) must hold since otherwise there exists  $w'\not \in \ell(w,v_L)$ that satisfies (ii). It follows that more than one feasible indirect match  to $v_L$ exists. Hence, there is insufficient information to determine which of them is the correct one, and an indirect match of $w$ and $v_L$ cannot be determined.

We next prove that if conditions (i)-(iii) hold, then $w$ is an indirect match of $v_L$. From condition (i) it follows directly that $w$ cannot be directly matched to $v_L$.
From condition (ii) it follows  that $\ell(w,v_L)$ exists; hence $w$ is a possible match. It is left to show that $w$ is the only feasible match. From condition (iii) it follows that $w$ is the only feasible match to $v_L$ since any other match, $w'$, satisfies $w'\in \tilde\ell(w,v_L)$. 
\vspace{1cm}

When (i) 
holds, an indirect match \crr{to a labeled solo vertex $v_L$} can be computed in a straightforward manner by traversing
the graph backwards from $v_L$ \crr{and} checking whether a vertex $w$ that satisfies (ii) and (iii) exists. This is clearly time consuming.
Instead, we propose to store auxiliary data  in each vertex; this data, which can be efficiently computed online from  the vertex itself and its parents, makes it possible to directly compute an indirect  match, if one exists. We will also use this data later for  scheduling.

\vspace{0.2cm}
\noindent{\bf Auxiliary data for matching by elimination:}
We define~$w$ to be an {\em origin} of $v$ if (i) $w$ is a solo vertex;
  (ii) there exists  an unlabeled path $\tilde \ell(w,v)$ and (iii) $w$ is either a source  of the graph \cbb{({\em unlabeled origin})} or a labeled vertex \cbb{({\em labeled origin})}.
A labeled vertex is the origin of itself and has no unlabeled origins. 
The set of origins of $v$ consists of the set of \cbb{vertices -- each associated with a distinct target ID -- that} may represent the same target as $v$.
\crrr{Note that only a labeled origin of $v$ may be directly matched to $v$.}

We observe that a \crr{solo vertex $v$} may have an indirect match only if it has at least one unlabeled origin (otherwise it can only be directly matched).
Furthermore,  $v$ may have an indirect match only if just one of its parents has unlabeled origins (otherwise, the  unlabeled origins, one from each parent, do not satisfy (iii) of Claim~{\bf 1}). Hence, to compute whether an indirect match exists, it is sufficient to store in each vertex the number of its unlabeled origins,  denoted by $n_{\neg L}(v)$, and   the single parent that has unlabeled sources, if one exists, $p_{\leftarrow} (v)$ (set to zero if one does not exist). 
Let $P(v)$ be the set of parents of~$v$. Both $n_{\neg L}(v)$ (given by summing 
the number of unlabeled origins of $P(v)$) and $p_{\leftarrow} (v)$  can  be recursively defined as follows:
\begin{equation} 
n_{\neg L}(v)=
\begin{cases}
		1						& v\not \in V_L ~~\& ~~|P(v)|=0 \\
		0						& v \in V_L \\
		\sum_{v_i\in P(v)}n_{\neg L}(v_i)     & \text{otherwise.} 
  \end{cases}
	\label{eq:unlabeled}
\end{equation}
\begin{equation} 
p_{\leftarrow} (v)= 
	\begin{cases}
		u & \exists !\text{ }u\text{ }|\text{ }u\in P(v) \text{ } \& \text{ } n_{\neg L}(u)>0 \\
   	0       & \text{otherwise.}
  \end{cases}
\end{equation} 

Note that  
if a vertex $w$ is the  indirect match of  a solo vertex $p_{\leftarrow} (v)$, it is also the indirect match of $v$. Therefore, we can efficiently and  recursively compute the single candidate of an indirect match \cbb{of $v$}, $C(v)$:
\begin{equation}
C(v)= 
	\begin{cases}
			0 & n_{\neg L}(v)=0 \\
			{C}(p_{\leftarrow} (v)) 	& p_{\leftarrow} (v)\neq 0 ~~\& ~~ C(p_{\leftarrow} (v))\neq 0 \\
			{solo}(v) 	& \text{otherwise},
  \end{cases}
\end{equation}
where ${solo}(v)$ holds $v$ if $v$ is a solo vertex  and 0 otherwise.

Note that if $v$ is a compound vertex, it cannot have an indirect match; however, the value $C(v)$ contains the candidate indirect match for its \cb{descendants}.
After labeling
\cbb{$v$} and untangling the MSG (if such untangling is possible), the auxiliary data is recalculated to be $n_{\neg L}(v)=0$. This reflects  that ambiguities of this target prior to the labeling are no longer relevant for future ambiguities. 
After each labeling
and once the untangling is complete, $C$ must also be recalculated for all the vertices that were disconnected from the solo chain during the untangling process. Each of these vertices then propagates the updated value to all its descendants, who recalculate their own values accordingly.

\subsection{Scheduling}
\label{sec:scheduling}

\comment{
\begin{figure*}[t!]%
\begin{tabular}{cccc} \hspace{-0.3cm}
\includegraphics[width=0.5\columnwidth]{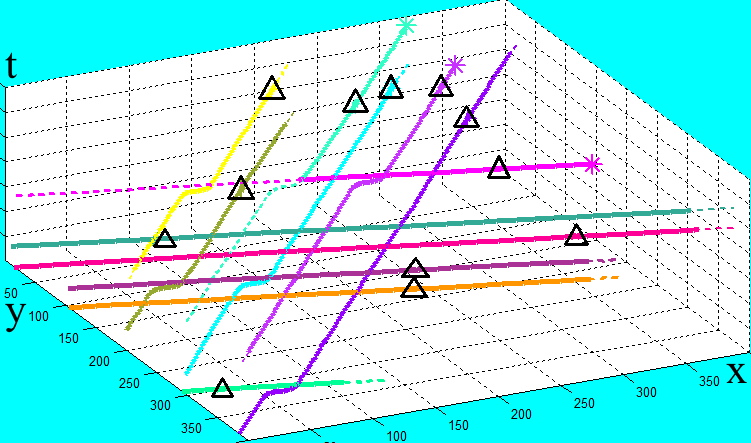}& \hspace{-0.3cm}
\includegraphics[width=0.5\columnwidth]{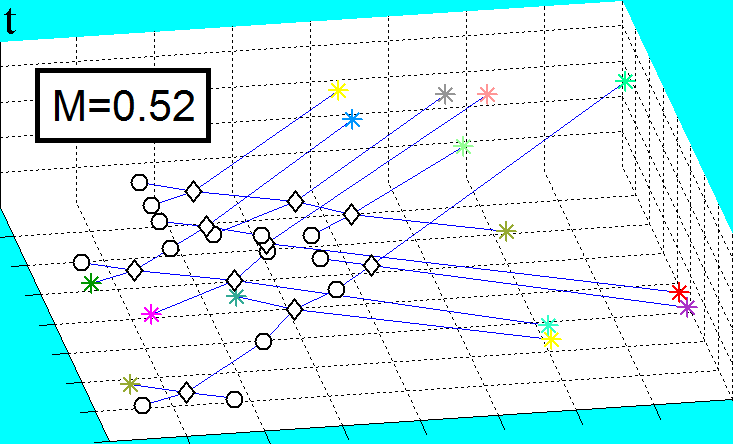}& \hspace{-0.3cm}
\includegraphics[width=0.5\columnwidth]{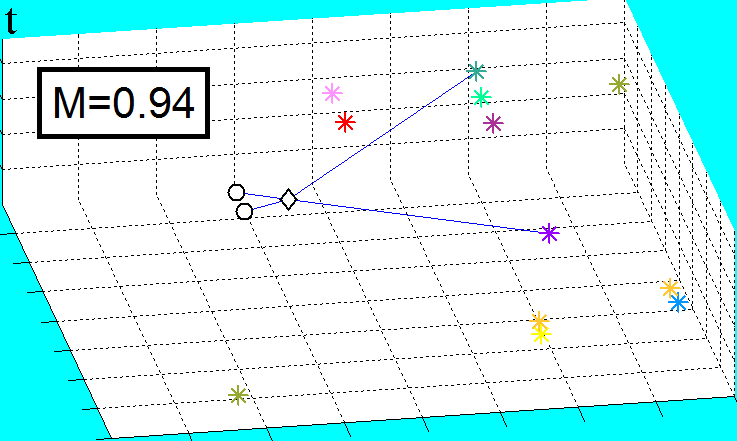}& \hspace{-0.3cm}
\includegraphics[width=0.5\columnwidth]{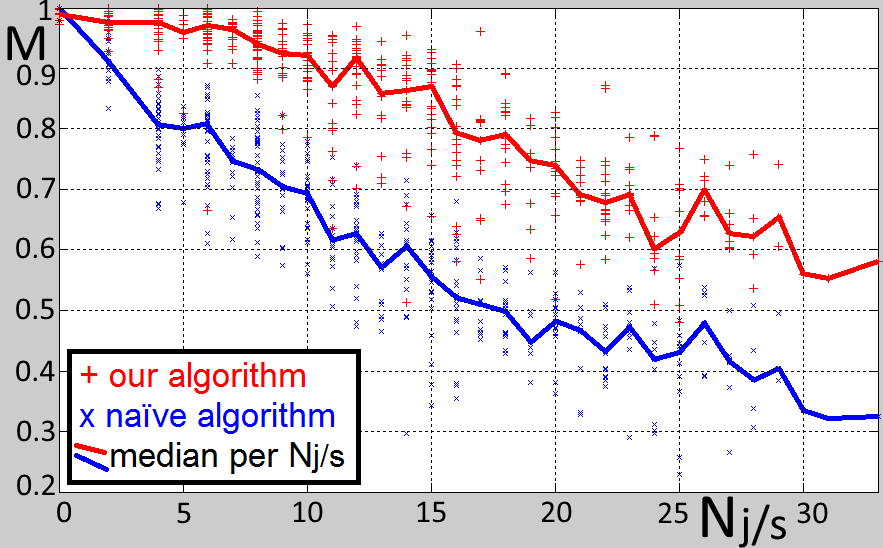}\\
(a) & (b) & (c) & (d)
\end{tabular}
\caption{Simulation screen shots. (a) \cb{Trajectory trails of all the targets that entered or already exited the scene, using (X,Y,t) coordinates. Solid lines: labeled tracklets; dashed lines: unlabeled tracklets; \cbb{black triangle: a face acquisition; asterisk: a labeled target.}} (b) The \cb{final MSG} \cbb{of the na\"{\i}ve scheduler.} (c) The \cb{final MSG} obtained by our method. Asterisk: a labeled vertex. (d) \cbb{$M$ as a function of $N_{j/s}$.} 
}%
\label{fig:simulation}%
\end{figure*}
}

\setlength{\fboxsep}{0pt}  
\setlength{\fboxrule}{2pt}  
\begin{figure*}[t!]%
\begin{tabular}{cccc} \hspace{-0.3cm}
\includegraphics[width=0.5\columnwidth]{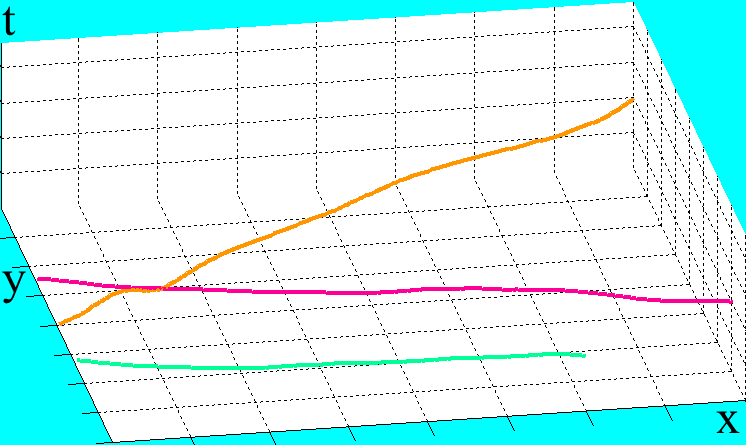}& \hspace{-0.3cm}
\fbox{\includegraphics[width=0.5\columnwidth]{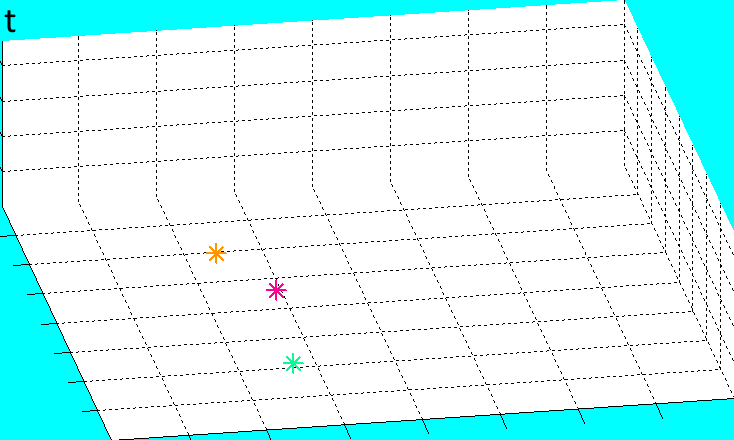}}& \hspace{-0.3cm}
\includegraphics[width=0.5\columnwidth]{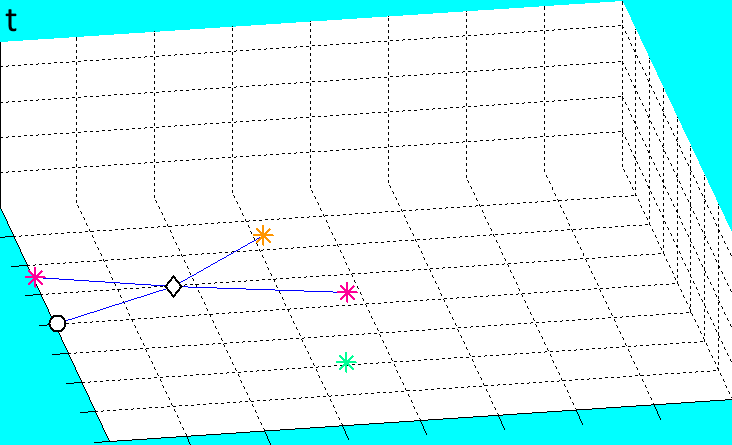}& \hspace{-0.3cm}
\includegraphics[width=0.5\columnwidth]{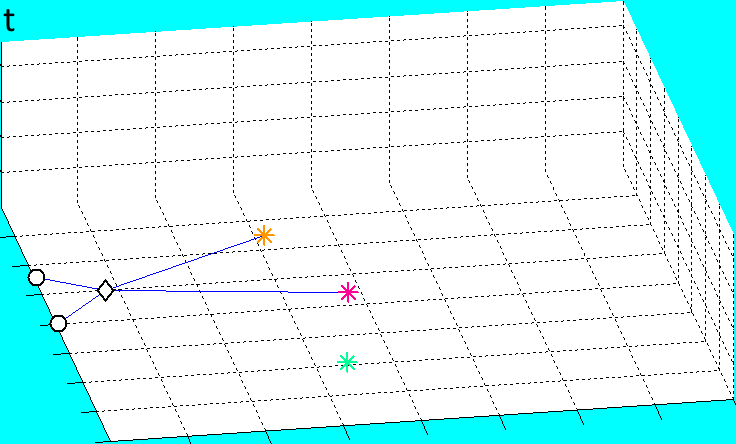}\\
(a) & (b) & (c) & (d)
\end{tabular}
\caption{\cnnn{Screen shots of simulation A. (a) Trajectory trails of 3 targets, using (X,Y,t) coordinates. (b) The final MSG obtained by our method. (c) The final MSG obtained by our method when untangling is not used. (d) The final MSG of the na\"{\i}ve scheduler. Asterisk: a labeled vertex.}} 
\label{fig:simple_simulation}%
\end{figure*}

The scheduler selects the tracklet   whose target's face will be acquired in the next zoom-in mode or selects to stay in zoom-out mode. The score \cbb{it provides} reflects the expected contribution of a tracklet labeling to maximize the system's objective, $M$ (Eq.~\ref{eq:M}). A prerequisite for choosing to \cbb{zoom in} on \cnn{an unlabeled} target is that  the acquisition of its face is expected to be successful.
A \cbb{Boolean} value that indicates  the expected success, $E_s(v)$, can be computed by the tracker in a similar manner to previous studies \cnn{(e.g., \cite{cai2013})}. For example,  the  motion direction can be used for predicting occlusions and time to \crr{exit}, and whether the face will be visible to the camera.

Ideally, the \cnn{minimal} number of required labelings for a full trajectory retrieval of $N$ targets is $N$,  one 
\cnn{per target}.
\cnn{The upper bound of the required number of ideal labelings }
is $\sum_i{(2n_i-1)}\leq2N-1$, \cnn{where $n_i=|G_i|$ and $G_i\subseteq G$ is a connected component.}
This sum includes the labeling of the first and last solo-walking tracklet of each target \crrr{$z$. Thus, the full trajectory of $z$} is recovered    and labeled
(under the  no-cycle assumption).  
\cbb{Note that each untangling may further reduce the required number of labelings.}

In practice, an ideal \crr{labeling set} is often impossible to obtain: \crr{the online algorithm leaves limited time}
for zooming in, and each labeling may cause additional ambiguities due to a blind gap. 
\crr{Moreover, the target identity} and hence its contribution to $M$ is unknown before zooming in. 

Therefore, we propose a scheduling algorithm \cbb{that approximates} the estimated contribution of labeling each of the targets or staying in  zoom-out mode to maximize $M$. 
A \emph{zoom-out score}, $S_{zo}$,  can reflect  global properties of the scene, such as the number of new targets expected to enter it, and the prevention of  X-type ambiguities  caused by a blind gap in zoom-in mode. Here we set it to be a constant. A \emph{labeling score}, $S_{L}(v)$, is set  for each vertex \cbb{and} reflects the   expected contribution to $M$ if $v$ is chosen to be labeled. 
The $v$ to be selected  for labeling  is the one with the highest $S_{L}$  as long as \cbb{$S_{L}(v)>S_{zo}$}. Otherwise,  the system remains in zoom-out mode.

\setlength{\fboxsep}{0pt}  
\setlength{\fboxrule}{2pt}  
\begin{figure*}[t!]%
\begin{center}
\begin{tabular}{c}
\begin{tabular}{ccc} \hspace{-0.6cm}
\includegraphics[width=0.7\columnwidth]{Figures/experiments_scene_3D}&  
\includegraphics[width=0.7\columnwidth]{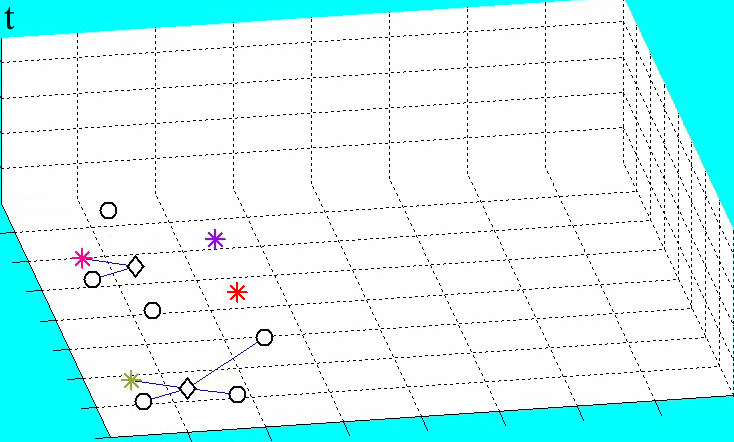}&  
\includegraphics[width=0.7\columnwidth]{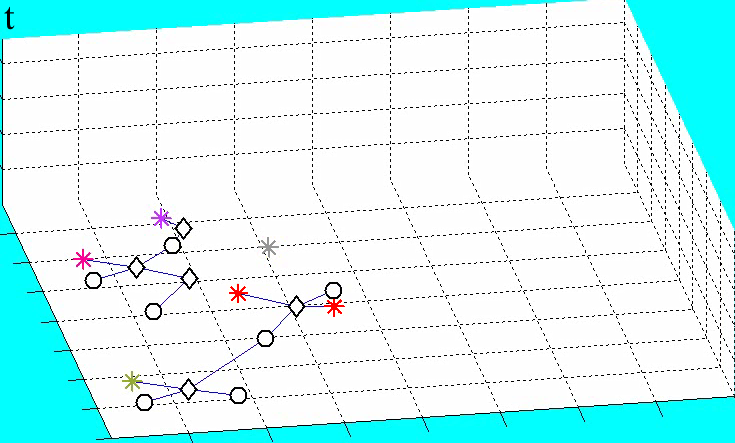}\\  \hspace{-0.6cm}
(a) & (b) & (c) \\ \\  \hspace{-0.6cm}
\includegraphics[width=0.7\columnwidth]{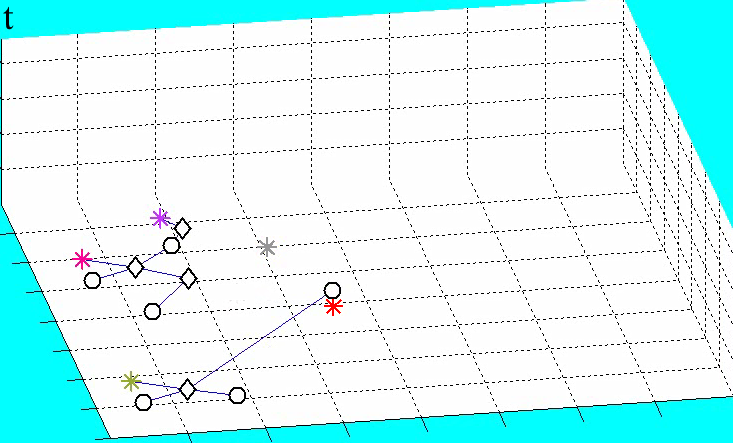} &  
\includegraphics[width=0.7\columnwidth]{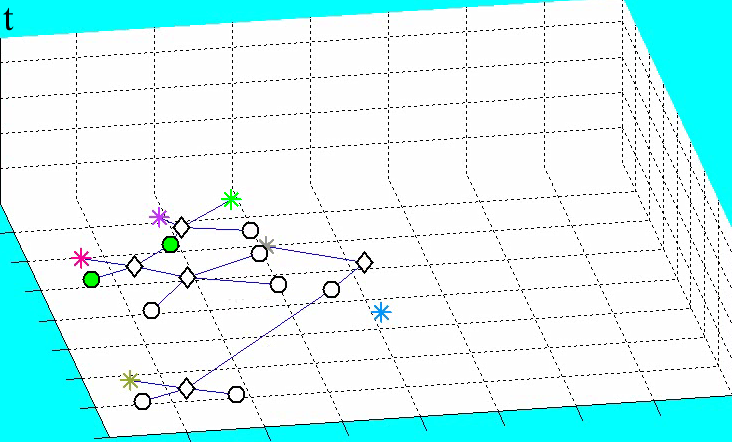}&   
\includegraphics[width=0.7\columnwidth]{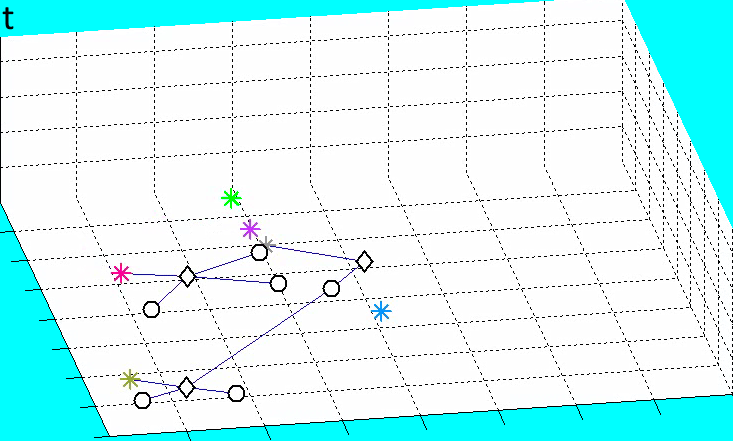}\\  \hspace{-0.6cm}
 (d) & (e) & (f) \\ \\
\end{tabular} \\
\begin{tabular}{cc} \hspace{-1.4cm}
\fbox{\includegraphics[width=0.8\columnwidth]{Figures/figure_3b_our_graph}}&   
\includegraphics[width=0.8\columnwidth]{Figures/figure_3b_naive_graph}\\ \hspace{-1.4cm}
  (g) & (h)
	\end{tabular}
	\end{tabular}
\end{center}
\caption{\cnnn{Screen shots of simulation B. (a) Trajectory trails of all the targets that entered or already exited the scene, using (X,Y,t) coordinates. Solid lines: labeled tracklets; dashed lines: unlabeled tracklets; black triangle: a face acquisition; asterisk: a labeled target. (b-f) Our scheduler's MSG after: (b) four targets are labeled, two of which before joining other targets; (c) a direct match event of the red vertices; (d) the following untangling; (e) an indirect match event of the green vertices; (f) the following untangling. (g) The final MSG obtained by our method. (h) The final MSG of the na\"{\i}ve scheduler. Asterisk: a labeled vertex.}} 
\label{fig:complex_simulation}
\end{figure*}

\subsection{Labeling Score \& Auxiliary Data}
\label{sec:labelingscore}
The score $S_{L}(v)$ is a weighted sum of two terms that estimate  the expected resolution of  future ($S_{F}$) and past ($S_{P}$) ambiguities: 
 \begin{equation}
\label{eq:Szi}
S_{L}(v)=E_s(v)\left(\alpha (v) S_{F}(v)+\beta (v){S}_{P}(v)\right),
\end{equation}
where the weights $\alpha (v)$ and $\beta (v)$ \cb{are} higher for source and expected sink vertices, respectively.

\vspace{0.1cm} \noindent{\bf Future ambiguities:}
The probability that $v$ was not labeled before is given by  $(n_{\neg L}(v)/n_o(v))$, where $n_o(v)$ and $n_{\neg L}(v)$ are  the number of  origins and unlabeled origins  of $v$, \cb{respectively}. The score  $S_F(v)$ is defined to be $
S_F(v)=Join(v){n_{\neg L}(v)/ n_o(v)}$,
where  $Join(v)$ is a \cbb{Boolean} value that reflects  that the target of $v$ is expected to join another target with a similar appearance (computed by the tracker).  
The value of  $n_{\neg L}(v)$  can be recursively computed (see Eq.~\ref{eq:unlabeled}). In a similar \crr{manner,} $n_o(v)$ can also be \cbb{recursively} computed \cnn{(as specified in Appendix A).}

\vspace{0.1cm} \noindent{\bf Past ambiguities:}
The score  $S_{P}(v)$ reflects the expected increase in the 
 length of the labeled tracklets, $L=\sum_{z\in Z}{|\tau_L(z)|}=\sum_{v\in V_L}{|\tau(v)|}$, if $v$ is chosen for labeling. Labeling a vertex $v$ increases the length of $L$ by the length of the tracklet $\tau(v)$. In addition, if $v$ is matched to $u$, directly or indirectly, then $L$ is extended by the sum of $|\tau(w)|$ over all $w\in \ell(u,v)$ such that $w\notin V_L$.
The identity of $v$'s target, and hence the origin to which~$v$ will be matched, is unknown prior to its labeling.  
 Hence, we  average over all possible \cbb{increases}  of $L$ with respect to the $n_{o}(v)$ possible origins that may be the \cbb{match} of $v$: 
\begin{equation}
S_{P}(v)={1\over n_{o}(v) } \left(\Delta L_{dir}(v)+L_{\neg dir}(v)\right),
\label{eq:deltaL}
\end{equation}
where $L_{\neg dir}(v)$ and $L_{dir}(v)$ 
 sum the increase of $L$ over the  sets of unlabeled origins  and labeled  origins, respectively.

A straightforward computation of $S_{P}(v)$ is by graph traversal.
To avoid such  a computationally  expensive operation for each candidate vertex,  we store in each vertex the auxiliary data fields, $\Delta L_{dir}(v)$, $L_{\neg dir}(v)$, and \crr{$n_o(v)$}. These values  are recursively computed. We next describe the computation of $L_{\neg dir}(v)$.
(For $\Delta L_{dir}(v)$, see Appendix~A.)

When no match is found (\crrr{either} direct or indirect), the  contribution of labeling $v$ to  $L_{\neg dir}(v)$ is  $|\tau(v)|$ for each unlabeled origin. 
  If an indirect candidate match is given by $C(v)$, that is,  $C(v)\notin \{0,v\}$, then the additional contribution of labeling $v$ 
is given by \crr{$|\tau(\ell (C(v),p_{\leftarrow} (v)))|$}. To compute it, $L_{\neg dir}(p_{\leftarrow} (v))$ is computed recursively as follows: 
{\small 
\begin{equation}
\begin{array}{l}
L_{\neg dir}(v)= \\ \\
	~~\begin{cases}
		|\tau (v)|\cdot n_{\neg L}(v)+L_{\neg dir}(p_{\leftarrow} (v))  & v\notin V_L ~~\&~~ p_{\leftarrow} (v)\neq 0\\
		|\tau (v)|\cdot n_{\neg L}(v)   & v\notin V_L ~~\&~~ p_{\leftarrow} (v)=0\\
		0       & v\in V_L.
	\end{cases}
\end{array}
\end{equation}}

\noindent{\bf Complexity:} The  computation of the score is linear with $|P(v)|$
\cbb{ -- which is expected to be small for each visible target -- instead of $O(|E|+|V|)$ for the necessary graph traversal without the auxiliary data.} Note that without 
untangling,
 the graph is expected to grow very fast when more targets enter the scene and many tracklets are detected, hence making the alternative $O(|E|+|V|)$ even worse. 
Due to the overhead incurred by untangling, the
auxiliary data of all \cbb{the} descendant vertices must be updated. In the worst case, it will require updating $O(|V|)$ vertices. However, this operation is rarely performed. Moreover, each time it takes place, the size of the graph is greatly reduced. 
Hence, the amortized complexity of updating the graph is expected to be $O(1)$ for each new vertex.  A formal proof of this conjecture  is left for future study.

\section{Experiments}
\label{sec:experiments}

\setlength{\fboxsep}{0pt}  
\setlength{\fboxrule}{2pt}  
\begin{figure*}[t!]%
\begin{center}
\begin{tabular}{cc} 
\includegraphics[width=0.8\columnwidth]{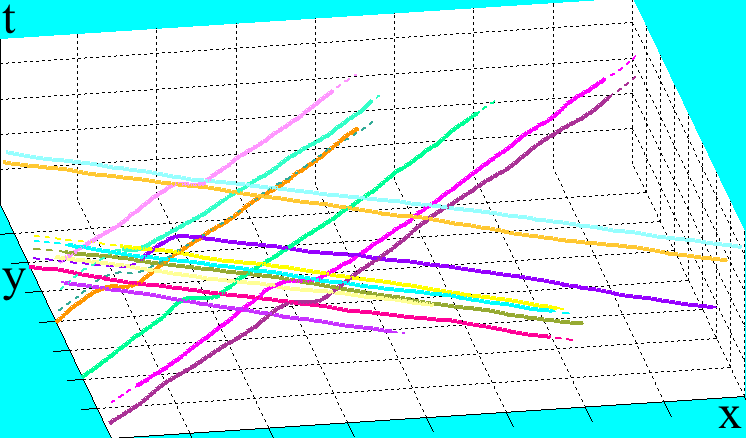}~~~~~~~ & ~~~~~~~
\fbox{\includegraphics[width=0.8\columnwidth]{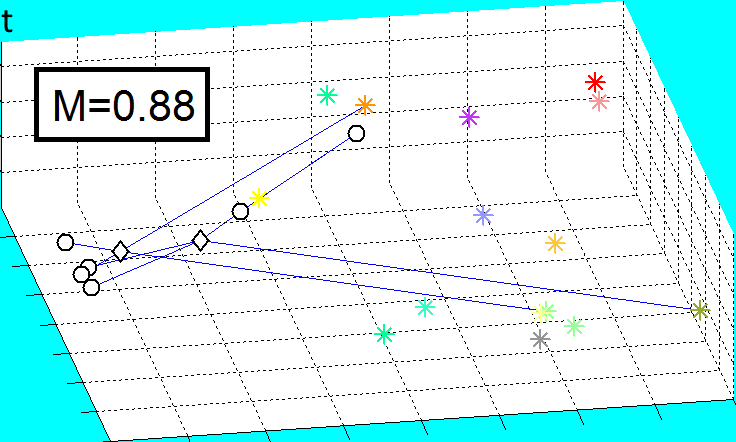}} \\
(a) & (b) \\ \\
\includegraphics[width=0.8\columnwidth]{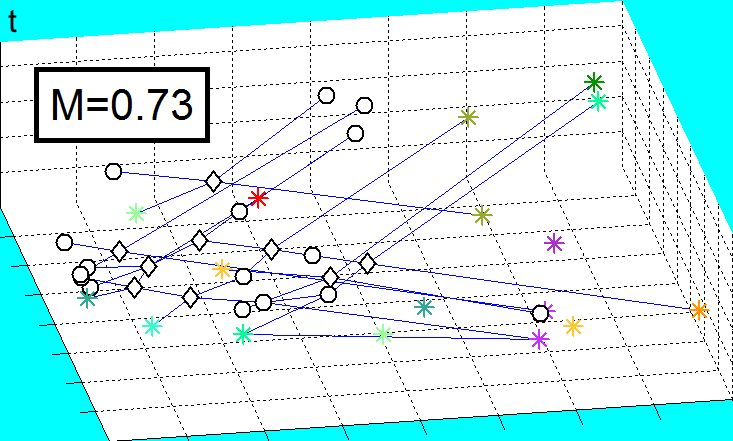}~~~~~~~ & ~~~~~~~ 
\includegraphics[width=0.8\columnwidth]{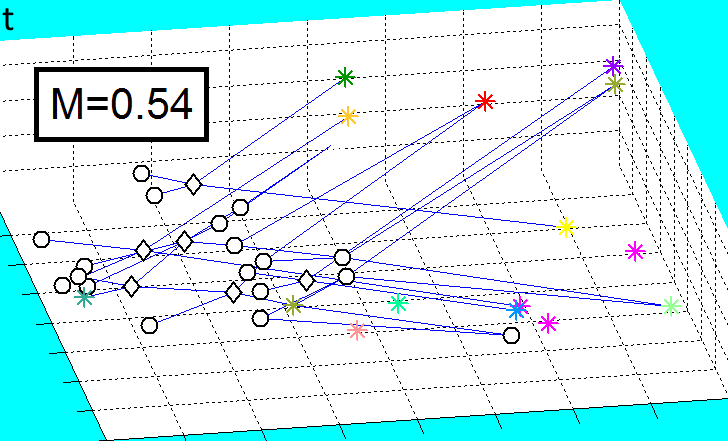} \\
(c) & (d)
\end{tabular}
\end{center}
\caption{\cnnn{Screen shots of simulation C. (a) Trajectory trails. (b) The final MSG of our method. (c) The final MSG of our method when untangling is not used. (d) The final MSG of the na\"{\i}ve scheduler. Asterisk: a labeled vertex.}} 
\label{fig:additional_simulation}%
\end{figure*}

 We used simulated data as an input to our method to evaluate the scheduler's performance independently from that of the other modules. Simulated data also 
\crrr{makes it possible} to bypass the limitations of comparing  online algorithms on the same  real data;
 each algorithm dictates different zoom-in operations, thus changing the data. We implemented our method as well as the simulated data using Matlab.

The  simulated scene  consists of a set of targets walking on a grid of intersecting diagonal roads. The targets' velocities (speed and direction), entrance time and location, \cbb{and the probability that meeting targets start walking together, are} determined randomly.
\cbb{All targets have} the same low-resolution appearance \cbb{to increase ambiguity}, \cnn{and low-resolution images are not used for matching.}

We present the objective score of our algorithm, $M$ (Eq.~\ref{eq:M}),  as a function of the expected ambiguities in the scene. \cnn{It is computed when the simulation ends and is based on the MSG's tracklets and on the ground truth.}
We estimate the ambiguities of the scene by  $N_{j/s}=\Sigma_{z\in Z} n_{j/s}(z)$, where $n_{j/s}(z)$ is the number of times a solo-walking target, $z$, joins a group and then splits to walk alone again. Note that in practice \cbb{blind gaps may cause additional ambiguities.}
For comparison we consider a na\"{\i}ve scheduler \cite{cai2013} that selects the unlabeled target predicted to leave the scene first. In both cases, we assume that the tracker provides the necessary available information (e.g.,  whether  the face is expected to be captured successfully).

\cnnn{A simple simulation of 3 targets and one join/split event (Figure~\ref{fig:simple_simulation}(a)) demonstrates a scenario where our scheduler selects the labeled target to be one of the two targets that are predicted by the tracker to join, before this event occurs. Consequently, one additional labeling after the split event untangles the MSG into an ideal graph, and all the trajectories are fully recovered (Figure~\ref{fig:simple_simulation}(b)). When our scheduler is used without the untangling process, its final MSG is not ideal and a full recovery is not achieved (Figure~\ref{fig:simple_simulation}(c)). The na\"{\i}ve scheduler selects the joining targets for labeling only after they split, thus preventing a full recovery and achieving the lowest $M$ (Figure~\ref{fig:simple_simulation}(d)). 
}
 
\cnnn{
Figure~\ref{fig:complex_simulation}(a) presents an example with a large number of targets and ambiguities. The MSG is growing rapidly but our scheduler achieves untangling in key points (see Figure~\ref{fig:complex_simulation}(b-f)), allowing the final MSG to contain only one remaining ambiguity (Figure~\ref{fig:complex_simulation}(g)). The na\"{\i}ve scheduler achieves a significantly lower $M$ due to a final MSG with many unresolved ambiguities (Figure~\ref{fig:complex_simulation}(h)). Another complex example is presented in Figure~\ref{fig:additional_simulation}.
}

\begin{figure}[t!]%
\begin{tabular}{c}
\includegraphics[width=0.95\columnwidth]{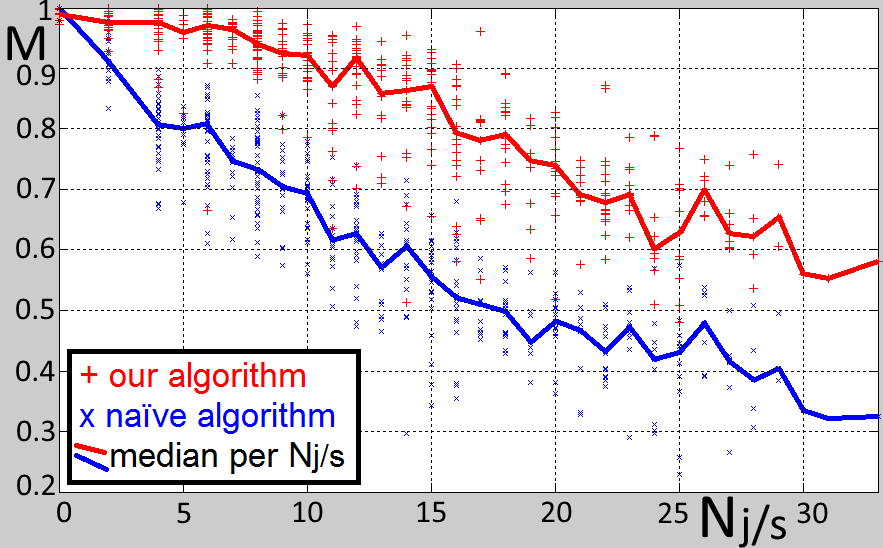}
\end{tabular}
\caption[]{\cnnn{Results of all simulations. $M$ as a function of $N_{j/s}$.}} 
\label{fig:results_of_all_simulations}%
\end{figure}

The results  on    414  simulations are  presented in Figure~\ref{fig:results_of_all_simulations}. When $N_{j/s}$ is small, the performance of  our  and the na\"{\i}ve algorithms is close to perfect. When $N_{j/s}$ increases, the performance of our method decreases, mainly due to the limited time available  to label \cb{all the} desired targets.
 However, for  moderate ambiguity of $N_{j/s}=15$,  our method  still performs well:   $M>0.85$. 
 The superiority of our algorithm over the na\"{\i}ve  \cb{one} is apparent both for moderate and high  $N_{j/s}$. For example, the score of the na\"{\i}ve  algorithm obtained for  $N_{j/s}=15$ is $M=0.55$, which is lower than the worst score of our method, for $N_{j/s}>30$.
	
Two components of our algorithm  contribute to its superiority over  the na\"{\i}ve  one. 
The global view we keep \cb{of the system state} allows us to associate  one or more labeled tracklets of the same target with  additional  tracklets of that target.  Using graph terminology, this corresponds to the untangling and merging of vertices, either by direct  labeling or as a byproduct of labeling other targets. 
In addition, our scheduling method explicitly considers the task of disambiguating tracklet \cb{associations}, and uses   global information of the current state of the system efficiently.

\section{Discussion \& Future Work}
\label{sec:summary}

We proposed a method for  tracking  multiple pedestrians and capturing their faces using a single PTZ camera. The goal of the system is to maximize the length of the labeled trajectories recovered by the \cbb{tracker}.  Our main contribution is a novel data structure, MSG,  that efficiently utilizes \cbb{all the available} global information of a tracking system. The auxiliary data of the MSG is used for an efficient scheduling algorithm  that resolves or  
\crr{prevents} tracklet ambiguities and  matches tracklets directly or indirectly 
\crr{via} target labeling. 
The MSG may be modified for various applications that use several cameras, with or without overlapping fields of view, when two distinct resolution levels can be used for resolving ambiguities. This is left for future research.

Our method aims to represent and efficiently use the data available from basic components of trackers and recognition systems, most of \cbb{which} are assumed to be deterministic for ease of exposition.  It is clearly prone to the expected errors of each of these components.

Our method  can be extended to handle a  probabilistic setting where  each  component provides a degree of confidence \cbb{for its} output. \cbb{This} can be integrated into the graph by,  for example, associating a weight with each edge.  In the current system,  X-type edges can represent the output of a  probabilistic person-to-person matching algorithm. A threshold on the face-to-face matching confidence may be used for deciding whether to untangle the graph or wait for additional information. 

\vspace{0.1cm} \noindent{\bf Acknowledgment:}~ This research was supported by the Israeli Ministry of Science, grant no. 3-8700, and by Award No. 2011-IJ-CX-K054, awarded by the National Institute of Justice, Office of Justice Programs, U.S. Department of Justice.

{\small
\bibliographystyle{ieee}
\bibliography{Shachaf-v8}

\begin{thebibliography}{10}\itemsep=-1pt

\bibitem{bagdanov2005}
A.~Bagdanov, A.~del Bimbo, and F.~Pernici.
\newblock Acquisition of high-resolution images through on-line saccade
  sequence planning.
\newblock In {\em VSSN}, 2005.

\bibitem{cai2013}
Y.~Cai, G.~Medioni, and T.~Dinh.
\newblock Towards a practical {PTZ} face detection and tracking system.
\newblock In {\em WACV}, 2013.

\bibitem{costello2004}
C.~Costello, C.~Diehl, A.~Banerjee, and H.~Fisher.
\newblock Scheduling an active camera to observe people.
\newblock In {\em VSSN}, 2004.

\bibitem{costello2005}
C.~Costello and I.~Wang.
\newblock Surveillance camera coordination through distributed scheduling.
\newblock In {\em CDCECC}, 2005.

\bibitem{henriques2011}
J.~Henriques, R.~Caseiro, and J.~Batista.
\newblock Globally optimal solution to multi-object tracking with merged
  measurements.
\newblock In {\em ICCV}, 2011.

\bibitem{lim2006}
S.~Lim, L.~Davis, and A.~Mittal.
\newblock Constructing task visibility intervals for video surveillance.
\newblock {\em MS}, 12(3), 2006.

\bibitem{morye2014}
A.~Morye, C.~Ding, A.~Roy-Chowdhury, and J.~Farrell.
\newblock Distributed constrained optimization for bayesian opportunistic
  visual sensing.
\newblock {\em TCST}, 2014.

\bibitem{Nillius2006}
P.~Nillius, J.~Sullivan, and S.~Carlsson.
\newblock Multi-target tracking-linking identities using bayesian network
  inference.
\newblock In {\em CVPR}, 2006.

\bibitem{prokaj2011}
J.~Prokaj, M.~Duchaineau, and G.~Medioni.
\newblock Inferring tracklets for multi-object tracking.
\newblock In {\em CVPRW}, 2011.

\bibitem{qureshi2007}
F.~Qureshi and D.~Terzopoulos.
\newblock Surveillance in virtual reality: System design and multi-camera
  control.
\newblock In {\em CVPR}, 2007.

\bibitem{salvagnini2014}
P.~Salvagnini, F.~Pernici, M.~Cristani, G.~Lisanti, I.~Masi, A.~Del~Bimbo, and
  V.~Murino.
\newblock Information theoretic sensor management for multi-target tracking
  with a single pan-tilt-zoom camera.
\newblock In {\em WACV}, 2014.

\bibitem{sommerlade2008}
E.~Sommerlade and I.~Reid.
\newblock Information-theoretic active scene exploration.
\newblock In {\em CVPR}, 2008.

\bibitem{arambel2004}
T.~Strat, P.~Arambel, M.~Antone, C.~Rago, and H.~Landan.
\newblock A multiple-hypothesis tracking of multiple ground targets from aerial
  video with dynamic sensor control.
\newblock In {\em SPIE}. 2004.

\bibitem{sullivan2006a}
J.~Sullivan and S.~Carlsson.
\newblock Tracking and labelling of interacting multiple targets.
\newblock In {\em ECCV}. 2006.

\bibitem{wang2014}
X.~Wang, E.~T{\"u}retken, F.~Fleuret, and P.~Fua.
\newblock Tracking interacting objects optimally using integer programming.
\newblock In {\em ECCV}. 2014.

\bibitem{ward2009}
C.~Ward and M.~Naish.
\newblock Scheduling active camera resources for multiple moving targets.
\newblock In {\em CCECE}, 2009.

\bibitem{Wu2011}
Z.~Wu, T.~Kunz, and M.~Betke.
\newblock Efficient track linking methods for track graphs using network-flow
  and set-cover techniques.
\newblock In {\em CVPR}, 2011.

\bibitem{yang2012}
B.~Yang and R.~Nevatia.
\newblock An online learned {CRF} model for multi-target tracking.
\newblock In {\em CVPR}, 2012.

\end{thebibliography}
}

\setcounter{equation}{7}

\section*{Appendix A}
This appendix provides the recursive computation of labeling score computation, $n_o(v)$ and $\Delta L_{dir}(v)$, 
 of \cbb{Section~\ref{sec:labelingscore}}.

\paragraph{Recursive computation of $n_o(v)$:}
The  number of origins of each vertex, $n_o(v)$, is recursively defined by:
\begin{equation} 	
n_o(v)=
\begin{cases}
		1						& v\in V_L \text{~~or~~} |P(v)|=0 \\
		\sum_{v_i\in P(v)}n_{o}(v_i)     & \text{otherwise}. 
  \end{cases}
	\label{eq:n_0}
\end{equation}
Note that using  $ n_{\neg L}(v)$ (given in  Eq.~2 of \cbb{Section~3.3}) and $n_o(v)$, we can also recursively compute the number of  labeled origins of~$v$:
\begin{equation} 
n_L(v)=n_o(v)-n_{\neg L}(v) \text{~~.}
\end{equation}

\paragraph{Recursive computation of $\Delta L_{dir}(v)$:}
Let us first consider the path $\ell(u,v)$, where $u$ is a  labeled origin of~$v$. Its contribution to $\Delta L_{dir}$ consists of $|\tau(\ell(u,v)|-|\tau(u)|$, since $u$ is a labeled vertex prior to the labeling of $v$.
Consider  $w\in \ell(u,v)\cap P(v)$, that is,  the parent of $v$ on the path $\ell(u,v)$. It is possible to decompose
$|\tau(\ell(u,v))|$ into the sum: $|\tau(\ell(u,v))|=|\tau(\ell(u,w))|+|\tau(v)|$. 
It follows that $v$ contributes $|\tau(v)|$ for each of its possible direct matchings, that is, $n_L\cdot|\tau(v)|$. In addition, the value $\Delta L_{dir}(v)$ consists  of the sum of $\Delta L_{dir}(w)$ for each of the parents of $v$ on possible direct match paths.
Hence, $\Delta L_{dir}(v)$ can be recursively computed: 
{\small \begin{equation}
	\Delta L_{dir}(v)=
	\begin{cases}
		|\tau (v)|\cdot n_L (v)+\sum^{}_{v_i\in P(v)}\Delta L_{dir}(v_i)  & v\notin V_L  \\ 
	    0  & v\in V_L.
		\end{cases} 
  \label{eq:Delta_L_dir}
\end{equation}
}

\vspace{-0.3cm}
\paragraph{\cbb{Refinement of the $\Delta L_{dir}(v)$ computation:}}
\cbb{The labeled tracklet of each labeled origin of $v$,  
$u$, clearly consists of the tracklet represented by $u$ itself, $\tau(u)$. 
In addition, a labeling of a vertex can sometimes be extended also to label  its parents and children. 
For example, assume that $v_1$ of a target $z$ is labeled in Figure~\ref{fig:ThreeTargets}(b).
The tracklet $\tau(v_4)$ clearly follows $\tau(v_1)$ for this target, and is therefore an extension of $\tau(v_1)$. 
Formally, let $v_L$ be a labeled vertex with a single compound child, $w$. The   tracklets  $\tau(v_L)$ and  $\tau(w)$ represent the same target. Hence,  $\tau(v_L)$ can be extended to $\tau(w)$. As a result, the length of the labeled tracklets is given by $|\tau(v_L)|+|\tau(w)|$. 
Such a {\em forward labeling extension} can be applied recursively to any {\em forward labeling chain},  $\sigma(v_L)$, which is a path from $v_L$ in which each vertex is a single child of its parent.
$\Delta L_{dir}(v)$ can be estimated more accurately
by considering forward labeling extensions  of the labeled origins of $v$, as described next.
}

\cbb{
Let us consider again the path $\ell(u,v)$, where $u$ is a  labeled origin of~$v$. We wish to find the contribution of $u$ to $\Delta L_{dir}(v)$ when considering not only $u$ itself but also its forward labeling extension. This contribution excludes the entire extension of $u$, which is labeled prior to the labeling of $v$. 
The length of the forward labeling extension of   $u$, $|\tau(\sigma(u))|$, is therefore subtracted from $|\tau(\ell(u,v))|$. That is, the contribution of the possible matching of $u$ to $v$ is given by 
$|\tau(\ell(u,v))|-|\tau(\sigma(u))|$.
We next describe the auxiliary data needed to compute the refined $\Delta L_{dir}(v)$ efficiently.
}

\cbb{
For each vertex $v$, we define the number of forward labeling chains in which $v$ is included, $n_{ret}(v)$. This value can be recursively computed based only on the vertex itself and its direct parents, as follows:
\begin{equation} 
	n_{ret}(v)=
	\begin{cases}
		\sum_{v_i\in P(v)}n_{ret}(v_i)\cdot {chain}(v_i)  & v\notin V_L \\ 
		1  & v\in V_L,
  \end{cases}
\end{equation} 
where the Boolean function ${chain}(v)$ determines whether $v$ \crr{has only one child}.
}
\cbb{
The refined recursive computation of $\Delta L_{dir}(v)$ (that replaces Eq.~\ref{eq:Delta_L_dir} above) is given by: 
{\small 
\[ 
\Delta L_{dir}(v)= ~~~~~~~~~~~~~~~~~~~~~~~~~~~~~~~~~~~~~~~~~~~~~~~~~~~~~~~~~~~~~~~~~~~~~~~~~~~~~~~~~~~~~~~~~
\]
\begin{equation}
	~~\begin{cases}
		|\tau (v)|\cdot (n_L(v)-n_{ret}(v))+\sum_{v_i\in P(v)}\Delta L_{dir}(v_i)  & v\notin V_L \\ 
		0  & v\in V_L.
  \end{cases}
\end{equation}}
}
\cbb{
Note that the labeling of a vertex can also be extended backwards, in a manner similar to the forward labeling extension. Both extensions are considered in the experiments for the evaluation of our scheduler, but only the forward labeling extension is useful for the refined $\Delta L_{dir}(v)$ computation.
}

\end{document}